\documentclass{article}

\usepackage{custom}
\usepackage{natbib}

\usepackage[utf8]{inputenc} 
\usepackage[T1]{fontenc}    
\usepackage{hyperref}       
\usepackage{url}            
\usepackage{booktabs}       
\usepackage{amsfonts}       
\usepackage{amsmath}
\usepackage{nicefrac}       
\usepackage{microtype}      
\usepackage{xcolor}         
\usepackage{graphicx}
\usepackage{multirow}
\usepackage{colortbl}
\usepackage{wrapfig}

\usepackage{algorithm,algpseudocode}

\usepackage{color}
\usepackage{xcolor}
\usepackage{soul}
\usepackage[normalem]{ulem}
\usepackage{booktabs}

\def\mypar#1{\vspace{0.15cm}\noindent{\bf #1.}}

\IfFileExists{darkmode.tex}{
    \usepackage{pagecolor}
    \definecolor{myfg}{gray}{0.94} 
    \definecolor{mybg}{gray}{0}
    \input{darkmode.tex} 
    \pagecolor{mybg}
    \color{myfg}
}{} 

\usepackage{amsmath,amssymb,amsbsy,xspace}

\def\<{\langle}
\def\>{\rangle}

\makeatletter
\DeclareRobustCommand\onedot{\futurelet\@let@token\@onedot}
\def\@onedot{\ifx\@let@token.\else.\null\fi\xspace}

\def\eg{\emph{e.g}\onedot} 
\def\ie{\emph{i.e}\onedot}

\makeatother

\newcommand{\X}{\mathbf{X}}
\newcommand{\until}{\mathbf{U}}

\newcommand{\red}[1]{{\color{red}  #1}}

\newcommand{\gc}[1]{} 
\newcommand{\eag}[1]{} 
\newcommand{\green}[1]{} 

\usepackage{pgfplots}
\pgfplotsset{
  compat=1.10,
}
\definecolor{exactColor}{RGB}{56,181,71}
\definecolor{correctColor}{RGB}{133, 246, 124}
\definecolor{incorrectColor}{RGB}{237, 151, 77}
\definecolor{invalidColor}{RGB}{253, 74, 74}
\definecolor{grayColor}{RGB}{160, 160, 160}

\title{Learning to Estimate System Specifications in Linear Temporal Logic using Transformers and Mamba}

\author{%
  İlker Işık \\
  Department of Computer Engineering \\
  Middle East Technical University \\
  Ankara, Turkey \\
  \texttt{ilker@ceng.metu.edu.tr} \\
  \AND
  Ebru Aydin Gol \\
  Microsoft \\
  İstanbul, Turkey \\
  \texttt{ebruaydingol@microsoft.com} \\
  \And
  Ramazan Gokberk Cinbis \\
  Department of Computer Engineering \\
  Middle East Technical University \\
  Ankara, Turkey \\
  \texttt{gcinbis@ceng.metu.edu.tr} \\
}

\begin{document}

\maketitle

\begin{abstract}
Temporal logic is a framework for representing and reasoning about propositions that evolve over time. It is commonly used for specifying requirements in various domains, including hardware and software systems, as well as robotics. Specification mining or formula generation involves extracting temporal logic formulae from system traces and has numerous applications, such as detecting bugs and improving interpretability. Although there has been a surge of deep learning-based methods for temporal logic satisfiability checking in recent years, the specification mining literature has been lagging behind in adopting deep learning methods despite their many advantages, such as scalability. In this paper, we introduce autoregressive models that can generate linear temporal logic formulae from traces, towards addressing the specification mining problem. We propose multiple architectures for this task: transformer encoder-decoder, decoder-only transformer, and Mamba, which is an emerging alternative to transformer models. Additionally, we devise a metric for quantifying the distinctiveness of the generated formulae and a straightforward algorithm to enforce the syntax constraints. Our experiments show that the proposed architectures yield promising results, generating correct and distinct formulae at a fraction of the compute cost needed for the combinatorial baseline.

\end{abstract}

\section{Introduction}
\label{sec:intro}

Linear temporal logic (LTL) is an extension of propositional logic that offers a symbolism for reasoning about how propositions change over time~\citep{Pnueli77}.
The primary application area of temporal logic is formal verification, for specifying requirements and verifying system behaviors~\citep{Clarke2018HandbookOM,Baier2008PrinciplesOM}.
Due to their expressiveness and similarity to natural language, temporal logics have become popular as a specification formalism in various fields, \eg dynamic systems \citep{belta2017}, robotics--especially in motion planning \citep{4154829,8263808,Fainekos2009TemporalLM,Sun2022MultiagentMP}, and biology \citep{Batt2005ValidationOQ}.

\begin{figure}[t]
  \centering
  \includegraphics[width=.92\textwidth]{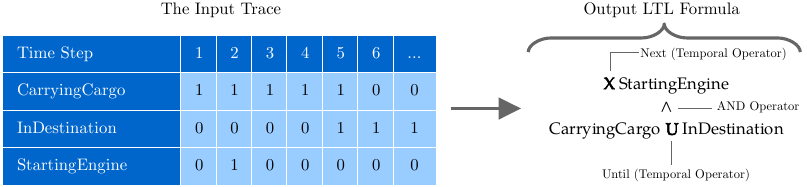}
  \caption{
      The overview of the problem.
      The table on the left denotes the input trace and shows how propositions evolve over time.
      On the right, a possible output is shown, which is a linear temporal logic (LTL) formula satisfied by the input trace.
      In this example, the formula dictates that the motor will start in the next time step, and the cargo will be carried until reaching the destination.
  }
  \label{fig:problem-overview}
\end{figure}

Extracting temporal logic formulae, typically as LTL or its variants, from system traces is the basis of specification (requirement) mining. Formulae extraction has many applications such as detecting bugs, testing for regressions, generating new tests, and so on~\citep{mineLtlSurvey,9465661,9110834,Bartocci2019AutomaticFE,Chanlatte2017,9096037}.
The resulting temporal logic formulae can also be used for the purposes of interpretability since LTL formulae are easily understood by human experts~\citep{mineLtlSurvey}.
Figure~\ref{fig:problem-overview} demonstrates the LTL specification mining problem.

The previous work in specification mining utilized a wide variety of methods, including template-based techniques~\citep{7084172}, methods based on decision trees~\citep{Bombara2016DecisionTree,8815002}, and many others~\citep{Chanlatte2017,Bartocci2014Data}.
These existing methods for specification mining either depend on human expertise (as in template-based methods) or suffer from combinatorial explosion problems.
In particular, the experiments by~\cite{Ghiorzi2023LearningLT} show that exhaustive combinatorial algorithms~\citep{syslite} and SAT-based solvers~\citep{ltlMaxsat} exhibit slow runtime performance, especially as the problem size grows, rendering them infeasible for practical applications.
Although \cite{Ghiorzi2023LearningLT} improved these baselines by devising clever optimizations that exploit the properties of LTL,
exhaustive combinatorial search scales poorly since the specification mining problem is NP-hard~\citep{ltlNP}.

%


Deep learning has attained a transformative success in computer vision~\citep{Liu2018DeepLF} and language modeling,
in which the large language models demonstrate impressive common sense reasoning abilities~\citep{Zhao2023ASO}.
Additionally, DeepLTL~\citep{deepltl} has exhibited excellent generalization performance in predicting traces from LTL formulae, showcasing that the deep learning models can learn the underlying semantics of LTL.
However, despite the success of deep learning in these fields,
there are no works that employ deep learning for specification mining to the best of our knowledge.
To overcome this deficiency in the specification mining literature, we propose several methods that are adapted from deep-learning-based natural language models.

The architectures we propose can be classified under two main categories.
The first one is based on the transformer encoder-decoder architecture~\citep{vaswani}, which has proved itself in natural language translation domain.
The second one is a series of decoder-only architectures, with the most prominent one being Mamba~\citep{mamba}. In our transformer encoder-decoder architecture, the specification mining is modeled as a
translation problem, with the source language being the input trace and the target language being
LTL.  Since both the input traces and LTL formulae are artificial languages, we use manually-defined
tokenizers unlike how it's done in natural language modeling.

The primary motivation for our decoder-only architectures is based on Mamba~\citep{mamba},
which reports promising improvements in a wide range of application areas, \eg language
modeling. To convert our problem into a sequence-to-sequence modeling problem, we merge the input
and output vocabularies.  The input trace is fed into the model and the model is expected to
generate the LTL formula autoregressively. In addition, we introduce a syntax-enforcing algorithm during the token generating phase to incorporate LTL syntax constraints.


We adapt the dataset from \cite{deepltl}, which was created by generating random LTL formulae and finding the satisfying traces. We introduce a distinctiveness metric that assesses the specificity of the generated formula in relation to the given trace. We present a detailed comparison of all our proposed models, including transformer, Mamba, and a decoder-only model based on Llama. The results show promising results on learning to generate semantically and syntactically correct, and distinct formulae, at a fraction of compute cost needed for combinatorial procedures.





\section{Related Work}
\label{sec:related}

\paragraph{Formula mining}
The problem of mining LTL formulae has been studied from different aspects in the literature,
such as by exploiting pre-existing SAT solvers~\citep{LTLSAT, ltlMaxsat} or automata~\citep{ltlAutomaton},
by using template-based algorithms~\citep{texada} or anytime algorithms~\citep{anytimeLtl},
and by using Bayesian inference to deliver formulae that contrasts multiple sets of traces~\citep{bayesianLtl}.

However, existing methods either encounter scalability issues and struggle to generate a formula in a timely manner, or make simplifying assumptions such as disregarding the until operator (as in \cite{anytimeLtl}).
The theoretical analysis by \cite{ltlNP} proved that learning LTL formulae from examples is a NP-hard problem.

\paragraph{Shift towards neural networks}

The introduction of neural networks to the temporal logic domain occurred through Signal Temporal Logic (STL), which is a variant of temporal logic that operates on continuous signals instead of propositions.
Specifically, \cite{stlcg} presented STLCG, which is a framework that defines computation graphs for the quantitative semantics of STL formulae, thereby enabling backpropagation through them.
This work depends on the fact that robustness metric for STL can be defined in a differentiable way.
Unlike LTL, the satisfaction of a STL formula is not binary; a robustness metric which denotes how well a signal fits a given formula can be calculated~\citep{stlRobustness}.
This enables numerical optimization methods that are not applicable for LTL.


\paragraph{Deep learning for temporal logic}
Although deep learning is not utilized for LTL formula mining,
several papers attacking other problems in temporal logic domain have been published.
Most importantly, \cite{deepltl} proposes a transformer encoder-decoder architecture to generate a satisfying trace for a given LTL formula, which is the opposite of the goal in formula mining, but still as hard as formula mining (PSPACE-complete as shown by \cite{complexLtlCheck}).

Another problem that captured attention within the temporal logic community was converting natural language statements to LTL formulae.
Reflecting the progress in language modeling literature, the first works that attack this problem utilized RNN encoder-decoder architecture~\citep{s2sLG,Patel2020GroundingLT}.
Then, the literature moved towards leveraging pre-trained large language models for better generalization abilities, pioneered by~\cite{langltl}.
Despite the fact that both this problem and formula mining involve extracting LTL formulae from another domain,
there is no prior work that employs deep learning methods for mining LTL formulae, to the best of our knowledge.

\paragraph{Language modeling}
The transformer architecture \citep{vaswani}, albeit initially developed for natural language translation, has been successfully applied to many other domains including genomic sequence modeling.
Mamba \citep{mamba} is a novel state-space model that emerged as an alternative to the transformer models with promising results in language modeling, a domain in which previous state-space models struggled.
In particular, Mamba language models matched the performance of transformer models twice their size and demonstrated better handling of long-distance relations.
We provide a more detailed summary of the Mamba architecture in the following section.


\section{Preliminaries}
\label{sec:prelim}

\subsection{Temporal logic overview}
\label{sec:prelim:temporal}

Linear Temporal Logic (LTL) is a superset of propositional logic that enables expressing and reasoning about how propositions change over time~\citep{Pnueli77}.
The LTL syntax over a finite set of atomic propositions $P$ is defined as follows:
\begin{equation}
  \phi := \mathbf{T} \mid p \mid \neg \phi \mid \phi_1 \wedge \phi_2 \mid \X \phi \mid \phi_1 \until \phi_2
\end{equation}
where $\mathbf{T}$ denotes \textit{True}, $p$ is an atomic proposition with $p \in P$, $\neg$ and $\wedge$ are the negation and conjunction operators respectively,
$\X$ and $\until$ are the temporal operators \textit{next} and \textit{until} respectively.
$\X \phi$ holds at time $t$ if and only if $\phi$ holds at time $t+1$.
$\phi_1 \until \phi_2$ implies that $\phi_2$ must hold immediately at the current time $t_1$ or at some point in the future $t_2$, and $\phi_1$ holds for all $t$ satisfying $t_1 \leq t < t_2$. For example, the formula $\X \X a$ requires $a$ to hold at the third time step. The formula $\mathbf{T} \until a$ requires $a$ to hold at some point in the future. Finally, the formula $\X b \wedge a \until c$ requires $b$ to hold at the second time step, c to hold at some point in the future and a to hold at all time points prior to that.

An LTL formula operates on a \textit{trace}, which is a sequence that denotes how the atomic propositions change over time.
As in DeepLTL \citep{deepltl}, we consider \textit{symbolic} traces of \textit{infinite} length, represented in a form known as ``\textit{lasso}''.
Such traces are denoted by $uv^\omega$, which comprises two sequences of propositional formulae: the prefix $u$ and the period $v$ that repeats infinitely. 
A symbolic trace represents all traces that satisfy these formulae at their respective time steps.  
For example, the symbolic trace $a, a \wedge \neg b, (c)^\omega$, represents all the traces in which $a$ holds during the first two time steps, $b$ does not hold at the second time step, and $c$ holds continuously from the third time step onward.
All of the traces represented by the symbolic trace, thus the symbolic trace itself, satisfy the formulae $\mathbf{T} \until c$ and $\X \neg b \wedge a \until c$. However, the symbolic trace violates $\X \X b$ as $b$ does not necessarily hold at the third time step.
Note that, as in this case, a symbolic trace can be underspecified, e.g., $a$ and $b$ can be in any configuration in the third time step.

\subsection{Problem definition}


Based on this background information, we can express our \textit{problem definition} as follows:
Given an input symbolic trace of the lasso form $uv^\omega$, find an LTL formula $\phi$ such that $uv^\omega$ satisfies $\phi$.
Moreover, it's desired that the generated formula is distinctive.
In particular, some trivial formulae such as $\mathbf{T}$ and $\neg (\neg a \wedge a)$ are satisfied regardless of the trace.
These formulae, albeit correct, are neither useful nor desirable since they don't convey any specific information about the traces they are supposed to describe.
We define a concrete performance metric for this concept in Section~\ref{sec:experiments:distinctiveness}.

\subsection{Language models}

\mypar{Transformers}
The goal of autoregressive language modeling is to predict the next token given the past tokens.
The transformer architecture~\cite{vaswani} uses attention layers based on a mechanism that predicts query, key and value tensors from inputs. In self-attention, these tensors are created come from the same inputs, while in cross-attention, key and value are predicted from different ones. The transformer encoder consists of self-attention and feed-forward layers, with the decoder adding cross-attention layers. Positional encodings are added to input embeddings to convey token order. Decoder outputs pass through a linear layer to produce logits, converted to probabilities via softmax. Attention masking during training ensures causality. 

\mypar{State Spaces \& Mamba} Mamba is a new type of state space model that has been proposed by \cite{mamba} as an alternative to transformers for sequence-to-sequence modeling.
A state space model is a continuous system \eqref{eq:ss} that uses the latent state $h(t) \in \mathbb{R}$ to map the 1-dimensional input $x(t) \in \mathbb{R}$ to the output $y(t) \in \mathbb{R}$,
where $\mathbf{A, B, C}$ are trainable model parameters.
The state space model can be discretized into \eqref{eq:ss-rnn}, introducing the $\mathbf{\overline{A}}$ and $\mathbf{\overline{B}}$ matrices, which provides RNN-like efficient inference over discrete input sequences \citep{lssm,s4}. The discrete model can also be converted into the convolutional form~\eqref{eq:ss-cnn}.

\vspace{-1em} 
\begin{minipage}{.30\textwidth}
  \begin{equation}
    \begin{gathered}
      h'(t) = \mathbf{A} h(t) + \mathbf{B} x(t) \\
      y(t) = \mathbf{C} h(t)
    \end{gathered}
    \label{eq:ss}
  \end{equation}
\end{minipage}
\begin{minipage}{.29\textwidth}
  \begin{equation}
    \begin{gathered}
      h_t = \mathbf{\overline{A}} h_{t-1} + \mathbf{\overline{B}} x_t \\
      y_t =  \mathbf{C} h_t
    \end{gathered}
    \label{eq:ss-rnn}
  \end{equation}
\end{minipage}
\begin{minipage}{.40\textwidth}
  \begin{equation}
    \begin{gathered}
      \mathbf{\overline{K}} =  (\mathbf{\overline{C}\overline{B}}, \mathbf{\overline{C}\overline{A}\overline{B}}, \mathbf{\overline{C}\overline{A}}^2\mathbf{\overline{B}}, \ldots) \\
      y = \mathbf{\overline{K}} \ast x
    \end{gathered}
    \label{eq:ss-cnn}
  \end{equation}
\end{minipage}

In Mamba, $\mathbf{B}$ and $\mathbf{C}$ matrices are derived from the input instead of being constant, which can be seen as a linear attention operator.
Furthermore, a Mamba block combines this sequence transformation with an MLP block that operates in the embedding space.
Similar to transformer, Mamba blocks are repeated to create the model, and the probabilities are computed using a linear layer and softmax after the last block.
Thanks to its recurrent nature, Mamba doesn't require any positional encoding.

\section{Proposed Method}
\label{sec:method}

\begin{figure}[t]
  \centering
  \includegraphics[width=\textwidth]{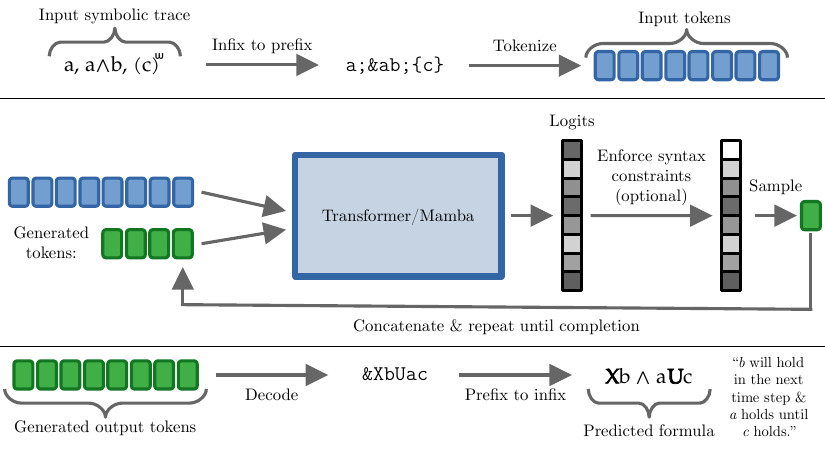}
  \caption{
      The visual summary of the proposed method.
      The input symbolic trace is converted into the prefix notation and then tokenized (Top).
      The model generates the formula tokens autoregressively, enforcing the syntax constraints if desired (Middle).
      The generated tokens are decoded and converted into the usual infix notation (Bottom).
  }
  \label{fig:summary}
\end{figure}

We tackle the LTL formula mining problem as an autoregressive language modeling task: given an input symbolic trace and the previous part of the generated LTL formula, the model predicts the next LTL token probabilities.
The following subsections explain the specialized tokenizer needed for traces and formulae, architectures, and syntax enforcing algorithm for incorporating LTL syntax constraints. Figure~\ref{fig:summary} provides a summary.

\subsection{Tokenizer}
\label{sec:method:tokenizer}

Our trace and formula syntaxes come from DeepLTL \citep{deepltl} for data compatibility. In both traces and formulae, we use the Polish notation, in which the operator is written first, e.g., $a \wedge b$ is denoted as \verb|&ab|.
This allows us to avoid grammar ambiguities without resorting to parentheses.

As explained in Section~\ref{sec:prelim:temporal}, we assume infinite symbolic traces of lasso form $uv^\omega$ in this work.
Alongside the characters used for atomic propositions, constants (\texttt{True:1} and \texttt{False:0}), and logical operators,
``\verb|;|'' character is used as a position delimiter, ``\verb|{|'' and ``\verb|}|'' mark the beginning and end of the period $v$.
For example, the string ``\verb|a;&ab;{b}|'' corresponds to the symbolic trace $a, a \wedge b, (b)^\omega$.

Since we work with a formal language with a predefined vocabulary, we define the tokenizer manually instead of training it.
Each character in the trace or formula vocabulary is assigned to a separate token.
Furthermore, special tokens are defined to mark the start or the end of the sequence, if needed.

\subsection{Encoder-Decoder and Decoder-only Architectures}
\label{sec:method:architectures}

Our model architectures can be grouped under two categories:
transformer encoder-decoder models and Mamba/Llama-based decoder-only models, as explained in the following.

\mypar{Transformer encoder-decoder} The first architecture we propose is a Transformer Encoder-Decoder from \cite{vaswani}.
The trace tokens and the previous formula tokens are fed into the encoder and decoder blocks respectively, with the cross-modal interactions between these two domains (\ie trace and formula) being handled by the cross-attention layers in the decoder.
The LTL formula is generated token-by-token by sampling from the predicted next token probabilities.
This architecture is essentially the same as the one in DeepLTL \citep{deepltl} with input and output swapped.


\mypar{Mamba} Mamba \citep{mamba} is a sequence-to-sequence model that doesn’t use any attention layers. Hence, the cross-attention mechanism which forms the backbone of transformer encoder-decoder is not applicable for Mamba.
Although achieving the effect of cross-attention using a Mamba architecture is still an open research question,
\cite{videomambasuite} discovered that a simple, direct concatenation of tokens from different modalities can
handle cross-modal interactions effectively, and thereby serve as a counterpart of cross-attention in Mamba-based models.

Based on this development by \cite{videomambasuite}, we rearranged our problem as a sequence-to-sequence problem by using concatenation in order to utilize Mamba.
Our Mamba model accepts both trace and formula tokens, but it only outputs the probabilities for the formula tokens.
During training, trace tokens are fed into the model as usual, but the model's predictions for these tokens are ignored by the loss function, as in the pad tokens.
Using this approach also removes the need for a start token.
Since the input trace is given in the same sequence, and ``\verb|}|'' token always marks the end of a trace and by extension, the beginning of the LTL formula.

Note that there are common tokens between the input and the output vocabularies, such as the atomic propositions, which may appear in both the input trace and the output formula.
Such tokens can either be shared by both traces and formulae, or be duplicated to create the trace and formula variants of the same token.
In this work, we duplicate such tokens. 


\begin{figure}[t]
  \centering
  \includegraphics[width=\textwidth]{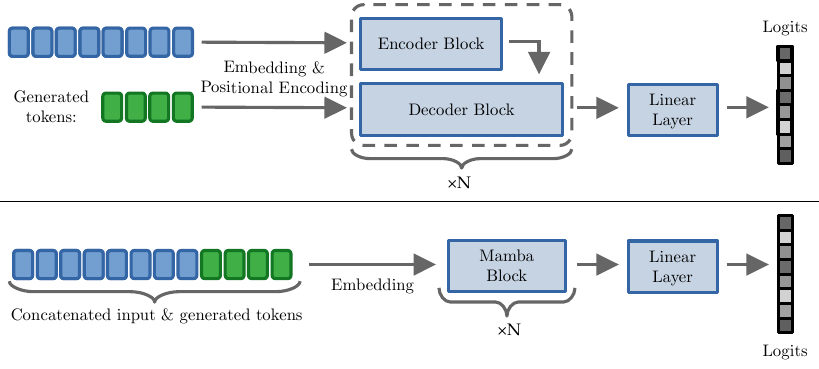}
  \caption{
    Model architectures, displaying the inner workings of the ``Neural Network Model'' block in Figure~\ref{fig:summary}.
    The transformer encoder-decoder model processes the tokens separately, using cross-attention mechanism to process the interactions between them (Top).
    In Mamba architecture, the input tokens and the generated tokens are concatenated in a single grammar (Bottom).
  }
  \label{fig:models}
\end{figure}

\mypar{Llama} To compare the Mamba model to a more similar transformer-based model, we additionally define a decoder-only transformer model based on Llama \citep{llama}.
The purpose of this model is to check whether switching to a decoder-only model (as we did for Mamba) improves the performance compared to a vanilla transformer encoder-decoder.
Furthermore, as noted by~\cite{mamba}, Llama is currently one of the strongest transformer recipes and outperforms the standard transformer architecture.
In addition, we also aim to observe whether the improvements in Llama, \eg RMSNorm instead of LayerNorm, SwiGLU activation, rotary embeddings, lead to a better model.

\subsection{Enforcing syntax constraints}
\label{sec:method:enforce}

Due to the nature of the sampling, the syntactic validity of the formula generated by the model is not guaranteed.
However, we devise a simple sampling method that enforces the syntax constraints.
The basic idea boils down to disallowing the tokens that would violate the syntactic rules, inspired from the method for generating JSON-formatted outputs in large language models~\citep{syntax-constrained}.


Thanks to the Polish notation, there are only two cases for syntactically invalid formula: either the formula ends prematurely (e.g., in \verb|Ua|, the right side of the binary operator $\until$ is missing), or the formula contains excess tokens (e.g., in \verb|Uabc|, $c$ does not belong to the parse tree).
Therefore, we can enforce syntactic validity just by controlling when it is allowed to emit the end token. Following this observation, 
we devise an algorithm that enforces the syntactic correctness by modifying the logits such that the generation of a syntactically invalid formula is impossible, assuming no limits on the generation length. The details are given in Appendix~\ref{sec:enforcing-exp}.

\section{Experiments}
\label{sec:experiments}

\subsection{Experimental setup}
\label{sec:experiments:setup}

We perfomed most of our experiments on a machine with Intel Core i9-10900X CPU @ 3.70GHz and NVIDIA RTX A4000, but we used different GPUs to train some of the models.
For evaluation, we make the models generate predictions for the symbolic traces, and then we use \verb|spot| framework version \verb|2.11.6| \citep{spot} to check the correctness of the predictions.
For each symbolic trace and LTL formula pair, we set the trace checking timeout to 30 seconds, which may be exceeded if the generated formula is too complex.
The timeout category also includes the pairs that caused a runtime error (e.g., overflow error) during trace checking.

\subsection{Quantitative semantic evaluation}
\label{sec:experiments:quantitative}

We trained multiple models on a preprocessed (Appendix~\ref{sec:hyperparam}) LTLRandom35 dataset from DeepLTL \citep{deepltl}, which consists of randomly generated LTL formulae and the corresponding symbolic traces.
Based on our hyperparameter analysis (Appendix~\ref{sec:hyperparam}), we determined the best performing model for each architecture and trained it three times from scratch in total.
We enumerate these models using letters A, B, C, and use the architecture name with the model letter to refer to them.
In Table~\ref{test-set-mean}, we give the average evaluation results of our these models on the test split of LTLRandom35 dataset, which contains 99038 samples.

\begin{table}[htbp]
    \caption{Mean and standard deviation of the test set results, over three runs. The best variant of each architecture has been selected on the validation set (the details can be found in the appendices).}
  \label{test-set-mean}
  \centering
    { 
  \begin{tabular}{lrrrrr}
    \toprule
    Architecture &  Correct $\%$ & Exact $\%$ & Incorrect $\%$ & Invalid $\%$ & Timeout $\%$ \\
    \midrule
    Transformer  & $94.38 \pm 4.12$ & $1.00 \pm 0.02$ & $4.45 \pm 3.68$ & $0.03 \pm 0.02$ & $1.14 \pm 0.50$ \\
    Mamba        & $98.23 \pm 0.29$ & $0.98 \pm 0.01$ & $1.38 \pm 0.11$ & $0.00 \pm 0.00$ & $0.39 \pm 0.18$ \\
    Llama        & $95.75 \pm 2.09$ & $0.93 \pm 0.01$ & $1.98 \pm 0.87$ & $0.00 \pm 0.00$ & $2.26 \pm 2.06$ \\
    \bottomrule
  \end{tabular} }
\end{table}

The most striking result is that although all models generate semantically correct formulae most of the time (Correct $\%$), they hardly ever generate the exact same formula (Exact $\%$) on the test set.
This shows that the models have learned the underlying semantics of the LTL formulae instead of memorizing and learning the specifics of the dataset.

\mypar{Comparison of architectures}
According to the average results in Table~\ref{test-set-mean}, the best architecture is Mamba, followed by Llama.
However, the full results (Table~\ref{test-set-full} in Appendix~\ref{sec:full-test})
reveal that the performance difference between the best models for each architecture is negligible.
Transformer and Llama models have higher variance in their performance, weakening their average results.
For a more detailed evaluation with all the models we trained, please see Appendices~\ref{sec:hyperparam} and \ref{sec:full-test}.

\mypar{The effect of syntax enforcing}
Before syntax enforcing, Transformer, Mamba and Llama models generated $(5.37 \pm 5.85)\%$, $(0.26 \pm 0.22)\%$, and $(6.76 \pm 5.23)\%$ invalid formulae respectively on the test set.
This shows that our Mamba models are better at complying with the LTL syntax without any external intervention.
Syntax enforcing converted $77\%$ of all invalid outputs on the test set into correct formulae.
More details are given in Appendix~\ref{sec:enforcing-exp}.

\subsection{Distinctiveness Evaluation}
\label{sec:experiments:distinctiveness}

The correctness of the generated LTL formulae is not the only desired quality.
The generated formulae should summarize the unique parts of the input trace to be helpful.
For instance, a trivial way to get $100\%$ semantic accuracy is to always generate the formula $\mathbf{T}$, which evaluates to \textit{True} regardless of the input trace and hence satisfies all traces.
To measure the network's tendency to generate such trivial formulae, we developed a distinctiveness metric.
Distinctiveness measure is based on a batch of symbolic traces and defined as in Eq.~\ref{eq:distinctiveness}:
\begin{equation}
  \label{eq:distinctiveness}
  1 - \frac{\textrm{Number of other traces that are satisfied}}{\textrm{Number of other traces}}
\end{equation}
Intuitively, the distinctiveness value is $0.0$ if the generated formula satisfies all other traces, as it's the case with $\mathbf{T}$.
Ideally, the distinctiveness value should be $1.0$, i.e., the generated formula should satisfy none of the other traces.
Note that the distinctiveness value is only applicable for correct formulae.

\begin{figure}[tbp]
  \noindent
  \begin{minipage}{.38\textwidth}
    \centering
    \includegraphics[width=\textwidth]{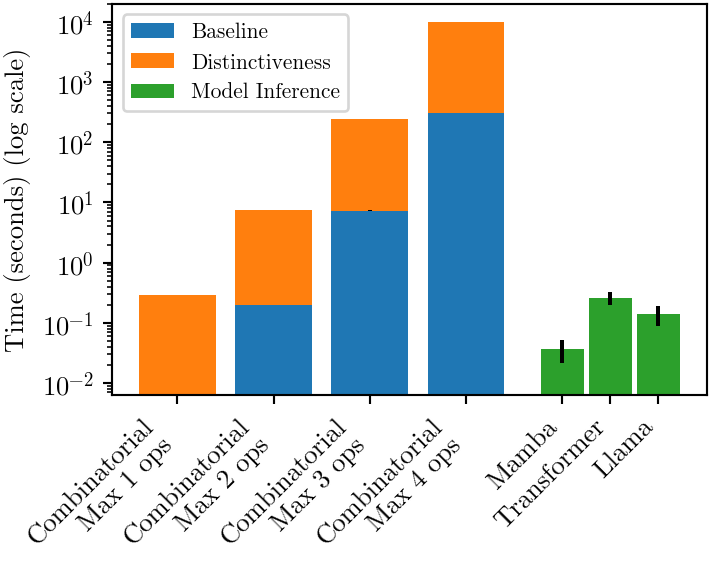}
    \caption{
     Execution times of the combinatorial algorithm with various operator count limits and our models.
     The optional distinctiveness computation checks each formula against 1000 traces to find the formula with best distinctiveness.
    }
    \label{fig:exectime}
  \end{minipage}%
  \hfill
  \begin{minipage}{.28\textwidth}
    \centering
    \includegraphics[width=\textwidth]{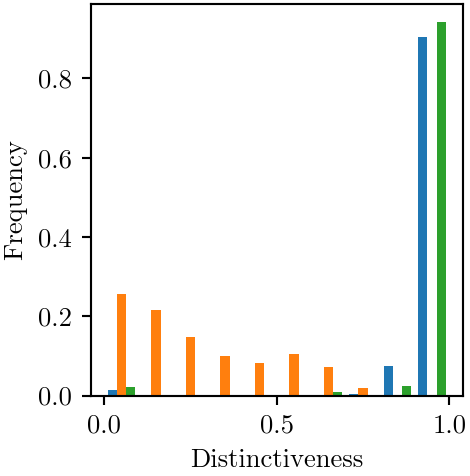} \vspace{0em}
          \caption{The distinctiveness distribution of the Mamba's predictions (blue) and the average/best (orange/green) of all formulae up to 4 operators. 
      }
    \label{fig:distinc-hist}
  \end{minipage}
  \hfill
  \begin{minipage}{.28\textwidth}
    \centering
    \includegraphics[width=\textwidth]{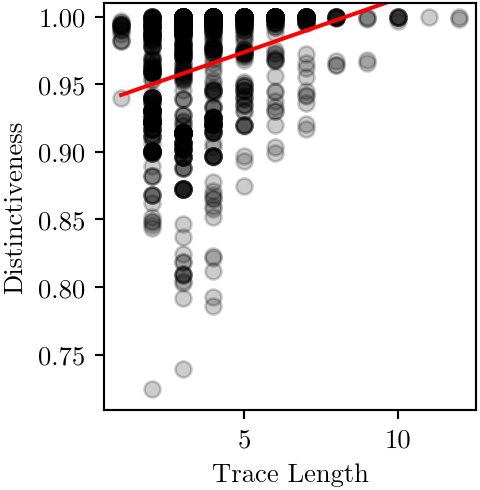}
    \caption{
      The scatter plot of the distinctiveness values of Mamba reported in Fig.~\ref{fig:distinc-hist} vs. the sequence length of the input symbolic trace (total length of $u$ and $v$).
      The line of best fit is shown in red. 
    }
    \label{fig:distinc-vs-tracelen}
  \end{minipage}
\end{figure}

Since the calculation of this metric requires checking $O(N^2)$ pairs,
we limit ourselves to a maximum of 1000 predictions to compute the distinctiveness values.
We calculated the distinctiveness values on a batch of 1000 samples from the test split of LTLRandom35.
Figure~\ref{fig:distinc-hist} visualizes the distinctiveness values for one of our models, Mamba.
As shown, the generated formulae have remarkably high distinctiveness values.
For this model, the average and the median of the distinctiveness values are $0.95$ and $0.99$ respectively, with $215$ formulae having the perfect distinctiveness value of $1.0$.
The other models perform similarly (Table~\ref{test-distinc-full} in Appendix~\ref{sec:full-distinc}).

There are only a few outlier cases in which the distinctiveness value is $0.0$.
However, upon closer inspection, the symbolic trace is just \verb|{1}| in these cases, which includes all possible traces.
This makes it impossible to come up with a unique formula.
Such degenerate cases are a byproduct of how the LTLRandom35 dataset was created.

Figure~\ref{fig:distinc-vs-tracelen} plots the distinctiveness values against the sequence length of the input trace, with the line of best fit shown in blue.
The figure demonstrates a positive correlation, i.e., the longer traces tend to be more specific, and as a result, the corresponding predictions have a high distinctiveness.
The same effect can be observed when comparing the values against the trace token count (Figure~\ref{mamba:distinc-vs-tracetokens} in Appendix~\ref{sec:full-distinc}).

\subsection{Combinatorial Baseline}
\label{sec:experiments:combinatorial}
In Figure~\ref{fig:exectime}, we compare the inference times of our models to a combinatorial approach that tests all possible formulae up to a certain operator count.\footnote{Rudimentary elimination rules (commutativity, associativity, double negatives, etc.) were implemented.}
Our Mamba model can create a formula $8327 \times$ faster than the combinatorial baseline (max 4 operators) with an average operator count of $17.74 \pm 11.33$. 
Note that the combinatorial approach outputs all formulae satisfying the input trace. As seen in Figure~\ref{fig:distinc-hist}, these formulae typically have low distinctiveness values in contrast to our models. Furthermore, we selected the formula with the highest distinctiveness value from those generated for the given trace and showed the results in Figure~\ref{fig:distinc-hist}. Although this method yields significantly high distinctiveness values, it also substantially increases the computation time (see Figure~\ref{fig:exectime}), to the extent that our Mamba model is $264019\times$ faster when distinctiveness analysis is integrated into the combinatorial approach. Note that our models demonstrate the capability to attain high distinctiveness values without integrating it into the selection or training criteria.

\subsection{Qualitative Evaluation}
\label{sec:experiments:qualitative}

In Table~\ref{tab:qualitative}, we inspect the predictions by all our models from the previous
section. In the first three rows, we show examples for three test traces where all nine models yield
syntactically the same semantically correct formulae.  There are $11120$ (out of $99038$) such
syntactically unanimous predictions in the test set, which suggests that similar system
specification priors are captured by drastically different architectures. 
Among these, in the first example, the ground truth (not shown in the table for brevity) is exactly the same as the prediction.

However, such exact matches to the ground truth are rare, as expected. In fact, we observe that the models can sometimes make syntactically simpler predictions compared to the ground truth, while being semantically correct. For example, in the example \#$2$ in Table~\ref{tab:qualitative}, 
the ground truth $\X (\neg (b \wedge
\neg \X e) \wedge  \X \X c \until  \X ((\neg d \until  d) \until  \X c))$ is clearly more complicated than the prediction.
Similarly, in the example \#$3$, the prediction is only a part of the ground truth $\neg d \wedge  \neg (1 \until  \neg b \wedge  \X ((1 \until  \neg c) \until  d \until  a))$.
Note that $\neg (1 \until  \neg b)$ means that $b$ holds in all time steps.
Applying De Morgan's law, the extra part in the ground truth becomes $\neg \X ((1 \until  \neg c) \until  d \until  a)$, disjuncted with the previous statement.
This statement, in addition to being unnecessarily complex with nested until statements, does not hold for the given trace, and consequently doesn't contribute to the formula descriptiveness.

%


For the following three traces in Table~\ref{tab:qualitative}, the models yield predictions that are not syntactically unanimous. For the trace \#$4$, 
the predicted formulas are semantically equivalent despite their syntactic differences.
\#$5$ and \#$6$ demonstrate the cases where the models produce subpar formula with
$\X$ operator repeats 30 and 25 times (denoted as $\X^{30}$ and $\X^{25}$), respectively. This tendency to generate semantically correct formulas with undesirable $\X$ repeats is an open problem for future work. 
%

\def\mytablerowsep{\arrayrulecolor{gray!60}\midrule\arrayrulecolor{black}}
\def\modelv#1#2{$\text{#1}_\text{#2}$}

\begin{table}[tbp]
  \caption{
    Example formula predictions for the trace samples from the LTLRandom35's test set. The predictions are semantically correct, unless denoted otherwise.
  \label{tab:qualitative}
  }
  \centering
\resizebox{\linewidth}{!}
    {
  \begin{tabular}{llcr}
    \toprule
    \# & Trace & Predicted formula  & Model \\
    \midrule 1 &

      $a \wedge  \neg e, e, d, (1)^\omega$ 
      & $a \until  e \wedge  \X \X d$ & All 9 models \\

    \mytablerowsep 2 &

      $1, b, d \wedge e, c, (1)^\omega$  
      & $\X (b \wedge  \X (d \wedge  e \wedge  \X c))$ & All 9 models \\

    \mytablerowsep 3 &

      $b \wedge  \neg d, (b)^\omega$ 
      & $\neg d \wedge  \neg (1 \until  \neg b)$ & All 9 models \\

    \mytablerowsep \multirow{3}{*}{4} &

    \multirow{3}{*}{$1, a \wedge  \neg c \wedge  \neg e, (a)^\omega$} 
      & $\X (\neg c \wedge  \neg e \wedge  \neg (1 \until  \neg a))$ & \modelv{Transformer}{A,B,C},\modelv{Mamba}{A} \\
    & & $\X \neg (1 \until  \neg a) \wedge  \X (\neg c \wedge  \neg e)$ & \modelv{Mamba}{B} \\
    & & $\X (\neg c \wedge  \neg e \wedge  \neg (1 \until  \neg a))$ & \modelv{Mamba}{C}, \modelv{Llama}{A,B,C} \\
    \mytablerowsep 5 &

    $1, \neg b, (1)^\omega$ 
    & $\neg (\X b \wedge  \X^{30} a)$ & \modelv{Transformer}{A} \\

    \mytablerowsep 6 &

    $1, b, e, (1)^\omega$ 
      & $\X b \wedge  \X \X (\neg \X^{25} b \until  e)$ & \modelv{Llama}{B} \\

    \mytablerowsep \multirow{2}{*}{7} &

    $\neg c, \neg c \wedge  \neg d, c \wedge  \neg d,$
    & $\neg (\neg c \until  \X d) \wedge  \X \X \X \neg (\X b \until  b)$ \red{\textit{(Incorrect)}} & \modelv{Transformer}{C},\modelv{Mamba}{B} \\
    & $\neg b, \neg b, (1)^\omega$
    & $\X \neg d \until  c \wedge  \neg \X \X \X (\X b \until  b)$ & \modelv{Transformer}{A} \\

    \mytablerowsep \multirow{2}{*}{8} &

    \multirow{2}{*}{$1, 1, c, (1)^\omega$} 
    & $\X \X (\X^{30} c \until \X c)$ \red{\textit{(Incorrect)}} & \modelv{Transformer}{B} \\
    & & $\X \X (\X^{25} \neg a \until  c)$ & \modelv{Mamba}{A} \\
    \bottomrule
  \end{tabular} }
\end{table}

The last two traces of Table~\ref{tab:qualitative} show semantically incorrect predictions.
In \#$7$, the prediction of \modelv{Transformer}{C} and \modelv{Mamba}{B} fails because some of the traces represented by the first symbolic trace satisfies $\neg c \until \X d$.
For instance, $d$ may hold at the fourth time step, and $\neg c$ holds until then; therefore,
the negation of $\neg c \until \X d$ is not satisfied due to such edge cases.
The \modelv{Transformer}{A} prediction, however, avoids this issue since it doesn't negate the until operator.
Finally, in the last trace, we again observe that the predictions contain ineffective until operators due to excessive $\X$s.
Here, \modelv{Transformer}{B}'s prediction fails due to an extra $\X$ before $c$, whereas the \modelv{Mamba}{A}'s prediction does not have this issue.

\section{Conclusion}
\label{sec:conclusion}

We have introduced an autoregressive generation approach for extracting LTL formulae from symbolic traces, addressing the lag in applying deep learning to specification mining.
We've proposed several different architectures for this task based on the ubiquitous transformer and the emerging Mamba, alongside a straightforward method to enforce LTL syntax constraints.
Our experimental results demonstrate the competency of our models in terms of semantic correctness and distinctiveness, which is measured using a metric we devised.
Furthermore, we have dissected the output formulae and discussed the qualitative limitations.
We believe the future work can improve our method by creating a better synthetic dataset for this task, 
and define new metrics for formula quality.

\begin{ack}
This study was supported by Scientific and Technological Research Council of Turkey (TUBITAK) under the Grant Number 122E249.
We thank to TUBITAK for their support.
We also gratefully acknowledge the computational resources kindly provided by METU Imagelab and METU-ROMER, Center for Robotics and Artificial Intelligence.
\end{ack}

{
\small
\bibliography{abbr,main}

\begin{thebibliography}{45}
\providecommand{\natexlab}[1]{#1}
\providecommand{\url}[1]{\texttt{#1}}
\expandafter\ifx\csname urlstyle\endcsname\relax
  \providecommand{\doi}[1]{doi: #1}\else
  \providecommand{\doi}{doi: \begingroup \urlstyle{rm}\Url}\fi

\bibitem[Arif et~al.(2020)Arif, Larraz, Echeverria, Reynolds, Chowdhury, and
  Tinelli]{syslite}
M.~F. Arif, D.~Larraz, M.~Echeverria, A.~Reynolds, O.~Chowdhury, and
  C.~Tinelli.
\newblock Syslite: Syntax-guided synthesis of pltl formulas from finite traces.
\newblock In \emph{2020 Formal Methods in Computer Aided Design (FMCAD)}, pages
  93--103, 2020.

\bibitem[Baier and Katoen(2008)]{Baier2008PrinciplesOM}
C.~Baier and J.-P. Katoen.
\newblock Principles of model checking.
\newblock 2008.

\bibitem[Bartocci et~al.(2014)Bartocci, Bortolussi, and
  Sanguinetti]{Bartocci2014Data}
E.~Bartocci, L.~Bortolussi, and G.~Sanguinetti.
\newblock Data-driven statistical learning of temporal logic properties.
\newblock In \emph{Formal Modeling and Analysis of Timed Systems}, pages
  23--37, Cham, 2014.

\bibitem[Bartocci et~al.(2019)Bartocci, Manjunath, Mariani, Mateis, and
  Ničković]{Bartocci2019AutomaticFE}
E.~Bartocci, N.~Manjunath, L.~Mariani, C.~Mateis, and D.~Ničković.
\newblock Automatic failure explanation in cps models.
\newblock In \emph{IEEE International Conference on Software Engineering and
  Formal Methods}, 2019.

\bibitem[Bartocci et~al.(2022)Bartocci, Mateis, Nesterini, and
  Nickovic]{mineLtlSurvey}
E.~Bartocci, C.~Mateis, E.~Nesterini, and D.~Nickovic.
\newblock Survey on mining signal temporal logic specifications.
\newblock \emph{Information and Computation}, 289:\penalty0 104957, 2022.
\newblock ISSN 0890-5401.

\bibitem[Batt et~al.(2005)Batt, Ropers, de~Jong, Geiselmann, Mateescu, Page,
  and Schneider]{Batt2005ValidationOQ}
G.~Batt, D.~Ropers, H.~de~Jong, J.~Geiselmann, R.~Mateescu, M.~Page, and
  D.~Schneider.
\newblock Validation of qualitative models of genetic regulatory networks by
  model checking: analysis of the nutritional stress response in escherichia
  coli.
\newblock \emph{Bioinformatics}, 21 Suppl 1:\penalty0 i19--28, 2005.

\bibitem[Belta et~al.(2017)Belta, Yordanov, and {Aydin Gol}]{belta2017}
C.~Belta, B.~Yordanov, and E.~{Aydin Gol}.
\newblock \emph{Formal Methods for Discrete-Time Dynamical Systems}.
\newblock Studies in Systems, Decision and Control. Springer, 2017.

\bibitem[Bombara et~al.(2016)Bombara, Vasile, Penedo, Yasuoka, and
  Belta]{Bombara2016DecisionTree}
G.~Bombara, C.-I. Vasile, F.~Penedo, H.~Yasuoka, and C.~Belta.
\newblock A decision tree approach to data classification using signal temporal
  logic.
\newblock pages 1--10, 04 2016.

\bibitem[Camacho and McIlraith(2021)]{ltlAutomaton}
A.~Camacho and S.~A. McIlraith.
\newblock Learning interpretable models expressed in linear temporal logic.
\newblock \emph{Proceedings of the International Conference on Automated
  Planning and Scheduling}, 29\penalty0 (1):\penalty0 621--630, May 2021.

\bibitem[Chen et~al.(2024)Chen, Huang, Xu, Pei, Chen, Li, Wang, Li, Lu, and
  Wang]{videomambasuite}
G.~Chen, Y.~Huang, J.~Xu, B.~Pei, Z.~Chen, Z.~Li, J.~Wang, K.~Li, T.~Lu, and
  L.~Wang.
\newblock Video mamba suite: State space model as a versatile alternative for
  video understanding, 2024.

\bibitem[Clarke et~al.(2018)Clarke, Henzinger, Veith, and
  Bloem]{Clarke2018HandbookOM}
E.~M. Clarke, T.~A. Henzinger, H.~Veith, and R.~Bloem.
\newblock Handbook of model checking.
\newblock In \emph{Cambridge International Law Journal}, 2018.

\bibitem[Duret-Lutz et~al.(2022)Duret-Lutz, Renault, Colange, Renkin, Aisse,
  Schlehuber-Caissier, Medioni, Martin, Dubois, Gillard, and Lauko]{spot}
A.~Duret-Lutz, E.~Renault, M.~Colange, F.~Renkin, A.~G. Aisse,
  P.~Schlehuber-Caissier, T.~Medioni, A.~Martin, J.~Dubois, C.~Gillard, and
  H.~Lauko.
\newblock From {S}pot 2.0 to {S}pot 2.10: What's new?
\newblock In \emph{Proceedings of the 34th International Conference on Computer
  Aided Verification (CAV'22)}, volume 13372 of \emph{Lecture Notes in Computer
  Science}, pages 174--187, Aug. 2022.

\bibitem[Fainekos et~al.(2009)Fainekos, Girard, Kress-Gazit, and
  Pappas]{Fainekos2009TemporalLM}
G.~Fainekos, A.~Girard, H.~Kress-Gazit, and G.~Pappas.
\newblock Temporal logic motion planning for dynamic robots.
\newblock \emph{Autom.}, 45:\penalty0 343--352, 2009.

\bibitem[Fijalkow and Lagarde(2021)]{ltlNP}
N.~Fijalkow and G.~Lagarde.
\newblock The complexity of learning linear temporal formulas from examples.
\newblock In \emph{Proceedings of the Fifteenth International Conference on
  Grammatical Inference}, volume 153 of \emph{Proceedings of Machine Learning
  Research}, pages 237--250, 23--27 Aug 2021.

\bibitem[Gaglione et~al.(2021)Gaglione, Neider, Roy, Topcu, and Xu]{ltlMaxsat}
J.~Gaglione, D.~Neider, R.~Roy, U.~Topcu, and Z.~Xu.
\newblock Learning linear temporal properties from noisy data: A maxsat-based
  approach.
\newblock In \emph{Automated Technology for Verification and Analysis - 19th
  International Symposium, ATVA 2021, Proceedings}, Lecture Notes in Computer
  Science (including subseries Lecture Notes in Artificial Intelligence and
  Lecture Notes in Bioinformatics), pages 74--90, Germany, 2021.

\bibitem[Ghiorzi et~al.(2023)Ghiorzi, Colledanchise, Piquet, Bernagozzi,
  Tacchella, and Natale]{Ghiorzi2023LearningLT}
E.~Ghiorzi, M.~Colledanchise, G.~Piquet, S.~Bernagozzi, A.~Tacchella, and
  L.~Natale.
\newblock Learning linear temporal properties for autonomous robotic systems.
\newblock \emph{IEEE Robotics and Automation Letters}, 8:\penalty0 2930--2937,
  2023.

\bibitem[Gopalan et~al.(2018)Gopalan, Arumugam, Wong, and Tellex]{s2sLG}
N.~Gopalan, D.~Arumugam, L.~L.~S. Wong, and S.~Tellex.
\newblock Sequence-to-sequence language grounding of non-markovian task
  specifications.
\newblock \emph{Robotics: Science and Systems XIV}, 2018.

\bibitem[Gu and Dao(2023)]{mamba}
A.~Gu and T.~Dao.
\newblock Mamba: Linear-time sequence modeling with selective state spaces.
\newblock \emph{arXiv preprint arXiv:2312.00752}, 2023.

\bibitem[Gu et~al.(2021)Gu, Johnson, Goel, Saab, Dao, Rudra, and R{\'e}]{lssm}
A.~Gu, I.~Johnson, K.~Goel, K.~Saab, T.~Dao, A.~Rudra, and C.~R{\'e}.
\newblock Combining recurrent, convolutional, and continuous-time models with
  linear state space layers.
\newblock \emph{{Advances in Neural Information Processing Systems}},
  34:\penalty0 572--585, 2021.

\bibitem[Gu et~al.(2022)Gu, Goel, and Ré]{s4}
A.~Gu, K.~Goel, and C.~Ré.
\newblock Efficiently {Modeling} {Long} {Sequences} with {Structured} {State}
  {Spaces}.
\newblock In \emph{Proc. Int. Conf. Learn. Represent.}, Aug. 2022.

\bibitem[Hahn et~al.(2021)Hahn, Schmitt, Kreber, Rabe, and Finkbeiner]{deepltl}
C.~Hahn, F.~Schmitt, J.~U. Kreber, M.~N. Rabe, and B.~Finkbeiner.
\newblock Teaching temporal logics to neural networks.
\newblock In \emph{9th International Conference on Learning Representations,
  {ICLR} 2021, Virtual Event, Austria, May 3-7, 2021}, 2021.

\bibitem[Jin et~al.(2015)Jin, Donzé, Deshmukh, and Seshia]{7084172}
X.~Jin, A.~Donzé, J.~V. Deshmukh, and S.~A. Seshia.
\newblock Mining requirements from closed-loop control models.
\newblock \emph{IEEE Transactions on Computer-Aided Design of Integrated
  Circuits and Systems}, 34\penalty0 (11):\penalty0 1704--1717, 2015.

\bibitem[Ketenci and Gol(2019)]{8815002}
A.~Ketenci and E.~A. Gol.
\newblock Synthesis of monitoring rules via data mining.
\newblock In \emph{2019 American Control Conference (ACC)}, pages 1684--1689,
  2019.

\bibitem[Kim et~al.(2019)Kim, Muise, Shah, Agarwal, and Shah]{bayesianLtl}
J.~Kim, C.~Muise, A.~Shah, S.~Agarwal, and J.~Shah.
\newblock Bayesian inference of linear temporal logic specifications for
  contrastive explanations.
\newblock In \emph{Proceedings of the Twenty-Eighth International Joint
  Conference on Artificial Intelligence, {IJCAI-19}}, pages 5591--5598, 7 2019.

\bibitem[Kloetzer and Belta(2007)]{4154829}
M.~Kloetzer and C.~Belta.
\newblock Temporal logic planning and control of robotic swarms by hierarchical
  abstractions.
\newblock \emph{IEEE Transactions on Robotics}, 23\penalty0 (2):\penalty0
  320--330, 2007.

\bibitem[Lemieux et~al.(2015)Lemieux, Park, and Beschastnikh]{texada}
C.~Lemieux, D.~Park, and I.~Beschastnikh.
\newblock General ltl specification mining (t).
\newblock In \emph{2015 30th IEEE/ACM International Conference on Automated
  Software Engineering (ASE)}, pages 81--92, 2015.

\bibitem[Leung et~al.(2023)Leung, Aréchiga, and Pavone]{stlcg}
K.~Leung, N.~Aréchiga, and M.~Pavone.
\newblock Backpropagation through signal temporal logic specifications:
  Infusing logical structure into gradient-based methods.
\newblock \emph{The International Journal of Robotics Research}, 42\penalty0
  (6):\penalty0 356--370, 2023.

\bibitem[Liu et~al.(2022)Liu, Yang, Schornstein, Liang, Idrees, Tellex, and
  Shah]{langltl}
J.~X. Liu, Z.~Yang, B.~Schornstein, S.~Liang, I.~Idrees, S.~Tellex, and
  A.~Shah.
\newblock Lang2{LTL}: Translating natural language commands to temporal
  specification with large language models.
\newblock In \emph{Workshop on Language and Robotics at CoRL 2022}, 2022.

\bibitem[Liu et~al.(2018)Liu, Ouyang, Wang, Fieguth, Chen, Liu, and
  Pietik{\"a}inen]{Liu2018DeepLF}
L.~Liu, W.~Ouyang, X.~Wang, P.~W. Fieguth, J.~Chen, X.~Liu, and
  M.~Pietik{\"a}inen.
\newblock Deep learning for generic object detection: A survey.
\newblock \emph{International Journal of Computer Vision}, 128:\penalty0 261 --
  318, 2018.

\bibitem[Mohammadinejad et~al.(2020)Mohammadinejad, Deshmukh, and
  Puranic]{9096037}
S.~Mohammadinejad, J.~V. Deshmukh, and A.~G. Puranic.
\newblock Mining environment assumptions for cyber-physical system models.
\newblock In \emph{2020 ACM/IEEE 11th International Conference on
  Cyber-Physical Systems (ICCPS)}, pages 87--97, 2020.

\bibitem[Neider and Gavran(2018)]{LTLSAT}
D.~Neider and I.~Gavran.
\newblock Learning linear temporal properties.
\newblock In \emph{2018 Formal Methods in Computer Aided Design (FMCAD)}, pages
  1--10, 2018.

\bibitem[Patel et~al.(2020)Patel, Pavlick, and Tellex]{Patel2020GroundingLT}
R.~Patel, E.~Pavlick, and S.~Tellex.
\newblock Grounding language to non-markovian tasks with no supervision of task
  specifications.
\newblock \emph{Robotics: Science and Systems XVI}, 2020.

\bibitem[Pnueli(1977)]{Pnueli77}
A.~Pnueli.
\newblock The temporal logic of programs.
\newblock In \emph{18th Annual Symposium on Foundations of Computer Science,
  Providence, Rhode Island, USA, 31 October - 1 November 1977}, pages 46--57,
  1977.

\bibitem[Puranic et~al.(2021)Puranic, Deshmukh, and Nikolaidis]{9465661}
A.~G. Puranic, J.~V. Deshmukh, and S.~Nikolaidis.
\newblock Learning from demonstrations using signal temporal logic in
  stochastic and continuous domains.
\newblock \emph{IEEE Robotics and Automation Letters}, 6\penalty0 (4):\penalty0
  6250--6257, 2021.

\bibitem[Raha et~al.(2022)Raha, Roy, Fijalkow, and Neider]{anytimeLtl}
R.~Raha, R.~Roy, N.~Fijalkow, and D.~Neider.
\newblock Scalable anytime algorithms for learning fragments of linear temporal
  logic.
\newblock In \emph{Tools and Algorithms for the Construction and Analysis of
  Systems}, pages 263--280, Cham, 2022.

\bibitem[Rehg(2023)]{syntax-constrained}
I.~Rehg.
\newblock Syntactically constrained sampling for language models.
\newblock \url{https://github.com/IsaacRe/Syntactically-Constrained-Sampling},
  2023.

\bibitem[Shoukry et~al.(2017)Shoukry, Nuzzo, Balkan, Saha,
  Sangiovanni-Vincentelli, Seshia, Pappas, and Tabuada]{8263808}
Y.~Shoukry, P.~Nuzzo, A.~Balkan, I.~Saha, A.~L. Sangiovanni-Vincentelli, S.~A.
  Seshia, G.~J. Pappas, and P.~Tabuada.
\newblock Linear temporal logic motion planning for teams of underactuated
  robots using satisfiability modulo convex programming.
\newblock In \emph{2017 IEEE 56th Annual Conference on Decision and Control
  (CDC)}, pages 1132--1137, 2017.

\bibitem[Sistla and Clarke(1982)]{complexLtlCheck}
A.~P. Sistla and E.~M. Clarke.
\newblock The complexity of propositional linear temporal logics.
\newblock In \emph{Symposium on the Theory of Computing}, 1982.

\bibitem[Sun et~al.(2022)Sun, Chen, Mitra, and Fan]{Sun2022MultiagentMP}
D.~Sun, J.~Chen, S.~Mitra, and C.~Fan.
\newblock Multi-agent motion planning from signal temporal logic
  specifications.
\newblock \emph{IEEE Robotics and Automation Letters}, PP:\penalty0 1--1, 2022.

\bibitem[Touvron et~al.(2023)Touvron, Lavril, Izacard, Martinet, Lachaux,
  Lacroix, Rozi{\`e}re, Goyal, Hambro, Azhar, Rodriguez, Joulin, Grave, and
  Lample]{llama}
H.~Touvron, T.~Lavril, G.~Izacard, X.~Martinet, M.-A. Lachaux, T.~Lacroix,
  B.~Rozi{\`e}re, N.~Goyal, E.~Hambro, F.~Azhar, A.~Rodriguez, A.~Joulin,
  E.~Grave, and G.~Lample.
\newblock Llama: Open and efficient foundation language models.
\newblock \emph{ArXiv}, abs/2302.13971, 2023.

\bibitem[Varnai and Dimarogonas(2020)]{stlRobustness}
P.~Varnai and D.~V. Dimarogonas.
\newblock On robustness metrics for learning stl tasks.
\newblock In \emph{2020 American Control Conference (ACC)}, pages 5394--5399,
  2020.

\bibitem[Vaswani et~al.(2017)Vaswani, Shazeer, Parmar, Uszkoreit, Jones, Gomez,
  Kaiser, and Polosukhin]{vaswani}
A.~Vaswani, N.~Shazeer, N.~Parmar, J.~Uszkoreit, L.~Jones, A.~N. Gomez, L.~u.
  Kaiser, and I.~Polosukhin.
\newblock Attention is all you need.
\newblock In \emph{Advances in Neural Information Processing Systems},
  volume~30, 2017.

\bibitem[Vazquez-Chanlatte et~al.(2017)Vazquez-Chanlatte, Deshmukh, Jin, and
  Seshia]{Chanlatte2017}
M.~Vazquez-Chanlatte, J.~Deshmukh, X.~Jin, and S.~Seshia.
\newblock Logical clustering and learning for time-series data.
\newblock pages 305--325, 07 2017.

\bibitem[Wang et~al.(2020)Wang, Cao, Tan, and Zong]{9110834}
F.~Wang, Z.~Cao, L.~Tan, and H.~Zong.
\newblock Survey on learning-based formal methods: Taxonomy, applications and
  possible future directions.
\newblock \emph{IEEE Access}, 8:\penalty0 108561--108578, 2020.

\bibitem[Zhao et~al.(2023)Zhao, Zhou, Li, Tang, Wang, Hou, Min, Zhang, Zhang,
  Dong, Du, Yang, Chen, Chen, Jiang, Ren, Li, Tang, Liu, Liu, Nie, and rong
  Wen]{Zhao2023ASO}
W.~X. Zhao, K.~Zhou, J.~Li, T.~Tang, X.~Wang, Y.~Hou, Y.~Min, B.~Zhang,
  J.~Zhang, Z.~Dong, Y.~Du, C.~Yang, Y.~Chen, Z.~Chen, J.~Jiang, R.~Ren, Y.~Li,
  X.~Tang, Z.~Liu, P.~Liu, J.~Nie, and J.~rong Wen.
\newblock A survey of large language models.
\newblock \emph{ArXiv}, abs/2303.18223, 2023.

\end{thebibliography}
}

\clearpage
\appendix

\section{Hyperparameter Analysis}
\label{sec:hyperparam}

In this section, we analyze the impact of different architectural hyperparameter choices.

\begin{figure}[htbp]
  \centering
  \includegraphics[width=.7\textwidth]{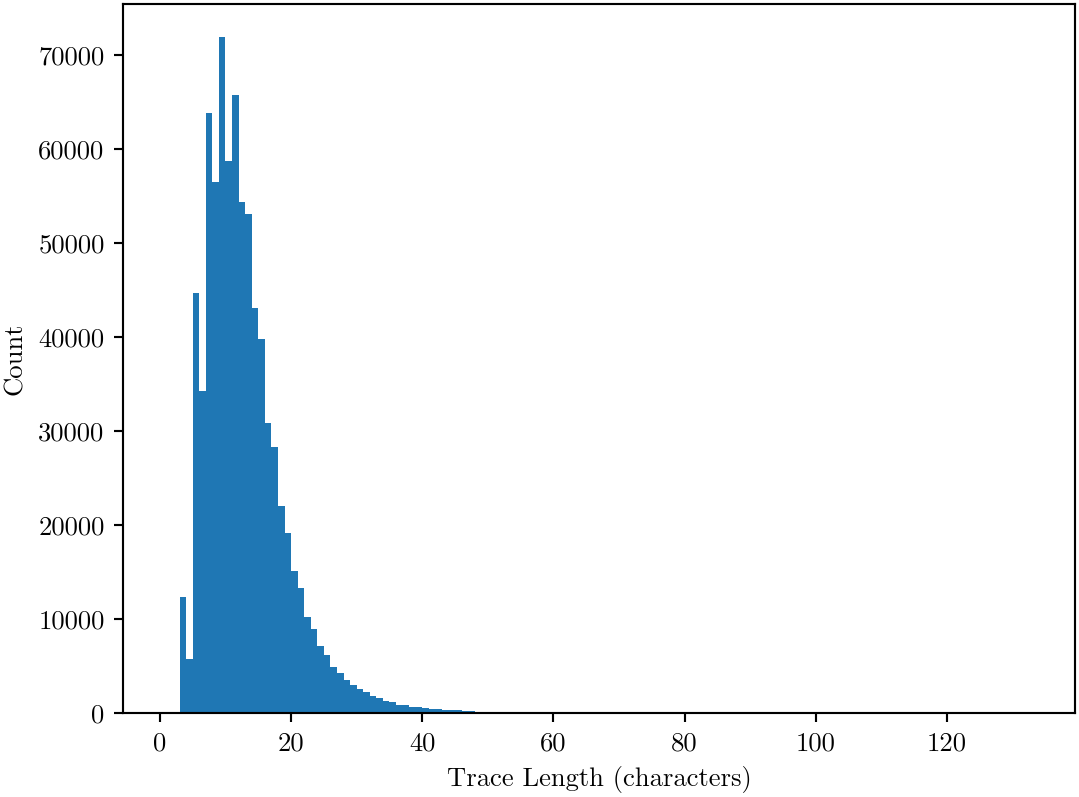}
  \caption{
    The distribution of the trace lengths in the training split of the LTLRandom35 dataset.
  }
  \label{tracelengthhist}
\end{figure}

We use the LTLRandom35 dataset from DeepLTL \citep{deepltl}, which contains pairs of randomly generated LTL formulae and the corresponding symbolic traces.
For preprocessing, we eliminate the pairs in which the length of the trace is longer than 35 characters in both training and evaluation.
Because the formula length distribution, as displayed in Figure~\ref{tracelengthhist}, reveals that such samples are exceedingly rare.

We trained each model type (transformer encoder-decoder, Mamba, Llama) 15 times with different hyperparameters.
We utilized grid search for our Mamba models since the Mamba language model presented by \cite{mamba} exposes only two model hyperparameters (dimension and layer count).
For other models, we searched more randomly since they have more hyperparameters to tune.

For all models, we used an AdamW optimizer with $\beta_1 = 0.9$, $\beta_2 = 0.95$, weight decay set to $0.1$, and gradient clipping set to $1.0$, as in Llama \citep{llama}.
We used a cosine learning rate schedule with 1000 warmup steps and a maximum learning rate of $0.001$.

\paragraph{Model naming conventions}
The model naming conventions are given below.
\begin{itemize}
  \item \textbf{Transformer and Llama-based:} \texttt{{ATTENTION\_HEADS}-{EMBED\_DIM}-{FF\_DIM}-{LAYERS}}
    \begin{itemize}
      \item \texttt{ATTENTION\_HEADS} is the number of attention heads in all transformer layers.
      \item \texttt{EMBED\_DIM} is the embedding dimension (for both encoder and decoder).
      \item \texttt{FF\_DIM} is the dimension of the feed-forward layers.
      \item \texttt{LAYERS} is the number of transformer blocks.
    \end{itemize}
  \item \textbf{Mamba:} \texttt{{DIMENSION}-{LAYERS}}
    \begin{itemize}
      \item \texttt{{DIMENSION}} is the model's embedding dimension.
      \item \texttt{{LAYERS}} is the number of Mamba blocks.
    \end{itemize}
  \item After the name, suffixes such as ``\texttt{-a}'' and ``\texttt{-b}'' can be added to denote the different training runs of the same architecture.
  \item The full names of the models mentioned in the main paper are as follows, where \verb|*| is \verb|a|, \verb|b|, or \verb|c| respectively:
    \begin{itemize}
      \item Transformer A, B, C: \verb|8-128-512-6-*|
      \item Mamba A, B, C: \verb|128-6-*|
      \item Llama A, B, C: \verb|8-64-128-6-*|
    \end{itemize}
\end{itemize}

During inference, we used beam search algorithm\footnote{For the transformer encoder-decoder and Mamba models, our beam search implementation is based on DeepLTL's \citep{deepltl} implementation, which in turn was adopted from TensorFlow. On the other hand, we used the HuggingFace's beam search implementation for Llama.} with a beam size of 3 in conjunction with the syntax enforcing.
We limited the generated formula length to 100 tokens, excluding the start token in the case of the transformer encoder-decoder architecture.
Please keep in mind that even though the syntax enforcing was enabled, the model can generate invalid formula if it exceeds the token limit.

Tables~\ref{ted-hyperparam}, \ref{mamba-hyperparam} and \ref{llama-hyperparam} show the evaluation results on the validation set.
Note that the samples classified as exact matches in ``Exact'' column are also included in the ``Correct'' column.
The ``Invalid'' column was omitted in Table~\ref{mamba-hyperparam} because none of the Mamba models generated any invalid formula.

\begin{table}[htbp]
  \caption{
    Evaluation of the transformer encoder-decoder models on 10000 validation set samples.
  }
  \label{ted-hyperparam}
  \centering
  \begin{tabular}{lrrrrrr}
    \toprule
    \multicolumn{2}{c}{Model} &
    \multicolumn{5}{c}{Evaluation} \\
    \cmidrule(r){1-2}
    \cmidrule(r){3-7}
    Name & Parameters & Correct & Exact & Incorrect & Invalid & Timeout \\
    \midrule
    4-32-64-2 & 44,142 & 6523 & 60 & 572 & 2877 & 28 \\
    4-32-64-4 & 86,894 & 9476 & 66 & 382 & 98 & 44 \\
    4-32-64-6 & 129,646 & 6643 & 72 & 243 & 3074 & 40 \\
    4-64-256-4 & 469,710 & 9766 & 74 & 117 & 0 & 117 \\
    4-128-512-4 & 1,856,910 & 9829 & 69 & 73 & 0 & 98 \\
    8-64-128-2 & 170,190 & 9674 & 62 & 284 & 0 & 42 \\
    8-64-128-4 & 337,614 & 9826 & 80 & 125 & 0 & 49 \\
    8-64-128-6 & 505,038 & 9748 & 75 & 106 & 2 & 144 \\
    8-64-256-2 & 236,238 & 9673 & 72 & 233 & 0 & 94 \\
    8-128-256-2 & 668,046 & 9696 & 73 & 150 & 44 & 110 \\
    8-128-256-8 & 2,655,630 & 9798 & 80 & 73 & 0 & 129 \\
    8-128-512-2 & 931,214 & 9782 & 76 & 118 & 2 & 98 \\
    8-128-512-6-a & 2,782,606 & 9833 & 79 & 59 & 0 & 108 \\
    8-128-512-6-b & 2,782,606 & 8895 & 78 & 907 & 4 & 194 \\
    8-128-512-6-c & 2,782,606 & 9547 & 78 & 378 & 1 & 74 \\
    8-128-512-8 & 3,708,302 & 9780 & 77 & 77 & 0 & 143 \\
    8-128-1024-8 & 5,813,646 & 9287 & 78 & 598 & 1 & 114 \\
    \bottomrule
  \end{tabular}
\end{table}

\begin{table}[htbp]
  \caption{
    Evaluation of the Mamba models on 10000 validation set samples.
    The ``Invalid'' column is omitted because it was all zeros.
  }
  \label{mamba-hyperparam}
  \centering
  \begin{tabular}{lrrrrr}
    \toprule
    \multicolumn{2}{c}{Model} &
    \multicolumn{4}{c}{Evaluation} \\
    \cmidrule(r){1-2}
    \cmidrule(r){3-6}
    Name & Parameters & Correct & Exact & Incorrect & Timeout \\
    \midrule
    32-2 & 20,960 & 7730 & 54 & 2241 & 29 \\
    32-4 & 40,864 & 9485 & 63 & 490 & 25 \\
    32-6-a & 60,768 & 9631 & 65 & 359 & 10 \\
    32-6-b & 60,768 & 9600 & 68 & 375 & 25 \\
    32-6-c & 60,768 & 9611 & 75 & 380 & 9 \\
    32-6-d & 60,768 & 9566 & 65 & 385 & 49 \\
    64-2 & 67,520 & 8981 & 61 & 997 & 22 \\
    64-4 & 132,928 & 9729 & 72 & 245 & 26 \\
    64-6 & 198,336 & 9788 & 76 & 180 & 32 \\
    128-2 & 237,440 & 9455 & 66 & 485 & 60 \\
    128-4 & 470,656 & 9797 & 81 & 154 & 49 \\
    128-6-a & 703,872 & 9843 & 91 & 127 & 30 \\
    128-6-b & 703,872 & 9769 & 93 & 146 & 85 \\
    128-6-c & 703,872 & 9836 & 85 & 137 & 27 \\
    256-2 & 884,480 & 9638 & 76 & 325 & 37 \\
    256-4 & 1,760,512 & 9736 & 76 & 197 & 67 \\
    256-6 & 2,636,544 & 9129 & 74 & 720 & 151 \\
    512-2 & 3,407,360 & 9600 & 82 & 392 & 8 \\
    512-4 & 6,797,824 & 9492 & 64 & 492 & 16 \\
    512-6 & 10,188,288 & 9501 & 71 & 486 & 13 \\
    \bottomrule
  \end{tabular}
\end{table}

\begin{table}[htbp]
  \caption{
    Evaluation of the Llama-based models on 10000 validation set samples.
  }
  \label{llama-hyperparam}
  \centering
  \begin{tabular}{lrrrrrr}
    \toprule
    \multicolumn{2}{c}{Model} &
    \multicolumn{5}{c}{Evaluation} \\
    \cmidrule(r){1-2}
    \cmidrule(r){3-7}
    Name & Parameters & Correct & Exact & Incorrect & Invalid & Timeout \\
    \midrule
    4-32-64-2 & 21,824 & 8656 & 51 & 1309 & 0 & 35 \\
    4-32-64-4 & 42,432 & 9388 & 65 & 555 & 0 & 57 \\
    4-32-64-6 & 63,040 & 9658 & 67 & 235 & 0 & 107 \\
    4-64-128-8 & 331,136 & 9155 & 74 & 415 & 0 & 430 \\
    4-128-512-4 & 1,054,464 & 8349 & 80 & 1108 & 0 & 543 \\
    8-64-128-2 & 84,608 & 9317 & 68 & 436 & 0 & 247 \\
    8-64-128-4 & 166,784 & 9672 & 66 & 154 & 0 & 174 \\
    8-64-128-6-a & 248,960 & 9815 & 76 & 111 & 0 & 74 \\
    8-64-128-6-b & 248,960 & 9583 & 70 & 338 & 0 & 79 \\
    8-64-128-6-c & 248,960 & 9273 & 72 & 199 & 0 & 528 \\
    8-64-128-8 & 331,136 & 9776 & 77 & 109 & 0 & 115 \\
    8-128-256-2 & 333,056 & 8998 & 63 & 891 & 0 & 111 \\
    8-128-256-8 & 1,317,632 & 9011 & 73 & 391 & 0 & 598 \\
    8-128-512-2 & 529,664 & 9284 & 67 & 594 & 0 & 122 \\
    8-128-512-4 & 1,054,464 & 9554 & 74 & 327 & 0 & 119 \\
    8-128-512-6 & 1,579,264 & 9430 & 73 & 290 & 0 & 280 \\
    8-256-512-6 & 3,944,960 & 8861 & 72 & 949 & 4 & 186 \\
    \bottomrule
  \end{tabular}
\end{table}

\section{Full Test Set Evaluation}
\label{sec:full-test}

In Table~\ref{test-set-full}, we give the full test set evaluation of the three training runs of the best models for each architecture,
which were reported as averages in Table~\ref{test-set-mean}.

\begin{table}[htbp]
  \caption{
    Full test set evaluation.
  }
  \label{test-set-full}
  \centering
  \begin{tabular}{llrrrrr}
    \toprule
    & & \multicolumn{5}{c}{Evaluation} \\
    \cmidrule(r){3-7}
    Architecture & Model Name & Correct & Exact & Incorrect & Invalid & Timeout \\
    \midrule
    \multirow{3}{*}{Transformer}
    & 8-128-512-6-a  &  97595 &    991 &    493 &      0 &    950 \\
    & 8-128-512-6-b  &  87921 &    979 &   9274 &     44 &   1799 \\
    & 8-128-512-6-c  &  94900 &   1015 &   3463 &     40 &    635 \\
    \midrule
    \multirow{3}{*}{Mamba}
    & 128-6-a        &  97530 &    981 &   1238 &      0 &    270 \\
    & 128-6-b        &  96881 &    961 &   1514 &      0 &    643 \\
    & 128-6-c        &  97434 &    958 &   1354 &      0 &    250 \\
    \midrule
    \multirow{3}{*}{Llama}
    & 8-64-128-6-a   &  97217 &    917 &   1031 &      0 &    790 \\
    & 8-64-128-6-b   &  95119 &    925 &   3109 &      0 &    810 \\
    & 8-64-128-6-c   &  92161 &    931 &   1756 &      0 &   5121 \\
    \bottomrule
  \end{tabular}
\end{table}
\section{Distinctiveness evaluation}
\label{sec:full-distinc}

\begin{figure}[htbp]
  \begin{minipage}{.48\textwidth}
    \centering
    \includegraphics[width=\textwidth]{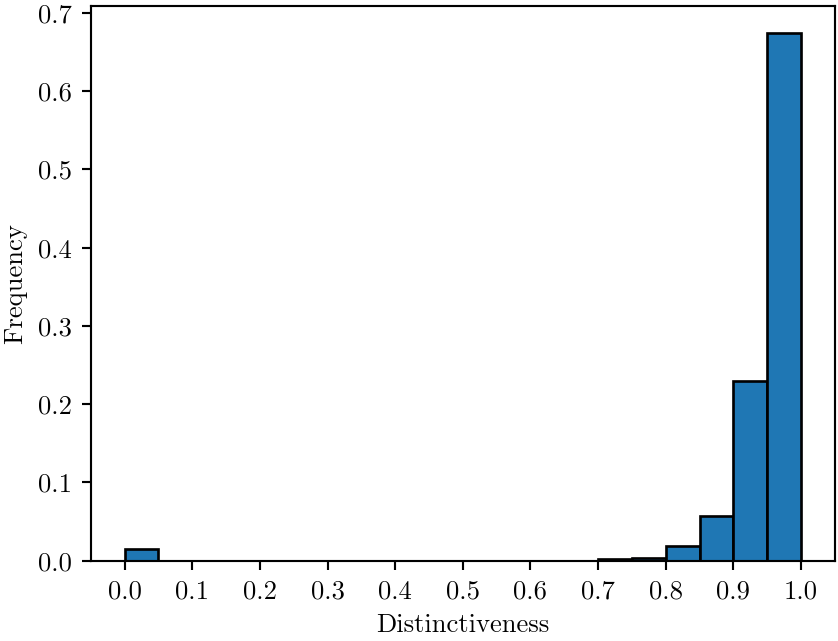}
    \caption{
      The histogram of the distinctiveness values of the
      formulae predicted by Mamba 128-6-a (Mamba A),
      calculated using 1000 test set pairs from LTLRandom35.
    }
    \label{mamba-a:distinc-hist}
  \end{minipage}%
  \hfill
  \begin{minipage}{.48\textwidth}
    \centering
    \includegraphics[width=\textwidth]{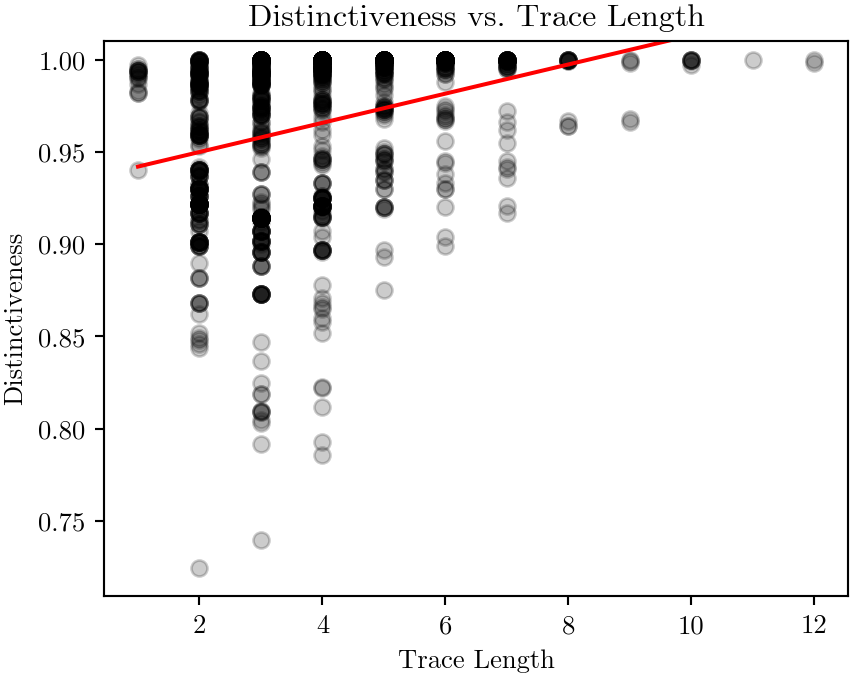}
    \caption{
      The scatter plot of the distinctiveness values reported in Fig.~\ref{mamba-a:distinc-hist} vs. the sequence length of the input symbolic trace (total length of $u$ and $v$).
      The line of best fit is shown in red.
    }
    \label{mamba-a:distinc-vs-tracelen}
  \end{minipage}
  \begin{minipage}{.48\textwidth}
    \centering
    \includegraphics[width=\textwidth]{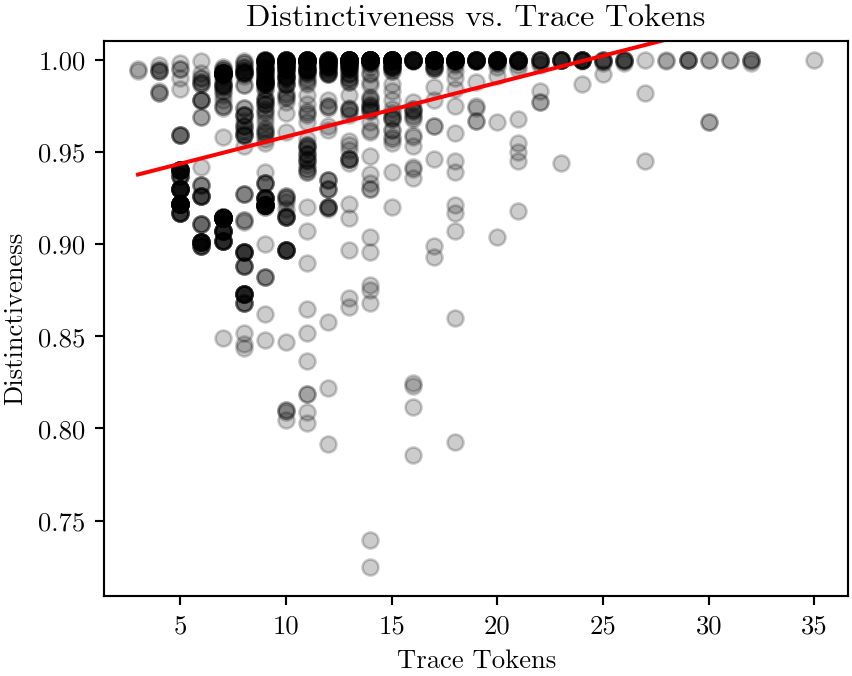}
    \caption{
      The scatter plot of the distinctiveness values reported in Fig.~\ref{mamba-a:distinc-hist} vs. the number of tokens in the input symbolic trace.
      The line of best fit is shown in red.
    }
    \label{mamba:distinc-vs-tracetokens}
  \end{minipage}%
  \hfill
  \begin{minipage}{.48\textwidth}
    \centering
    \includegraphics[width=\textwidth]{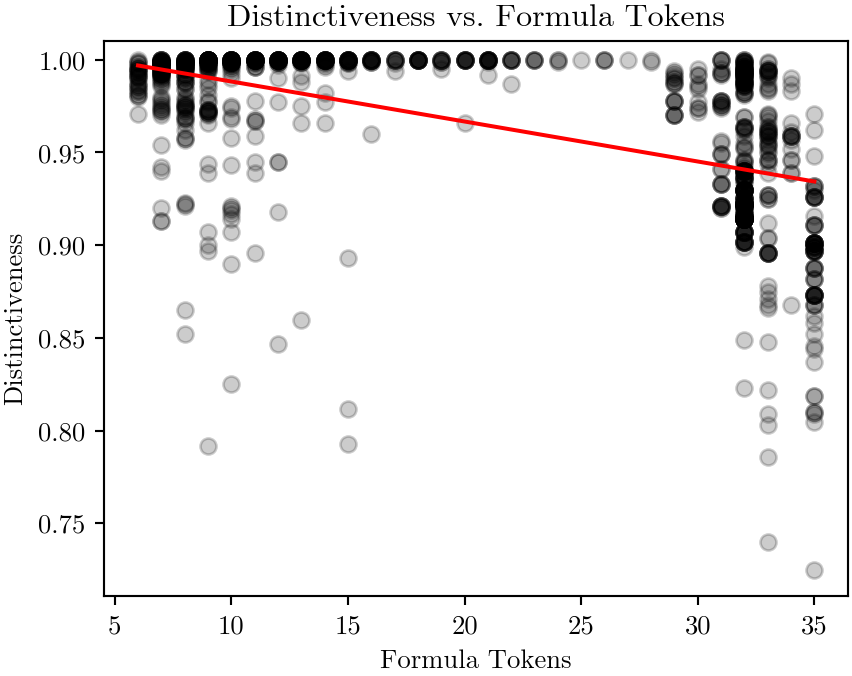}
    \caption{
      The scatter plot of the distinctiveness values reported in Fig.~\ref{mamba-a:distinc-hist} vs. the number of tokens in the generated output.
      The line of best fit is shown in red.
    }
    \label{mamba:distinc-vs-ltltokens}
  \end{minipage}
\end{figure}

\begin{figure}

  \begin{minipage}{.48\textwidth}
    \centering
    \includegraphics[width=\textwidth]{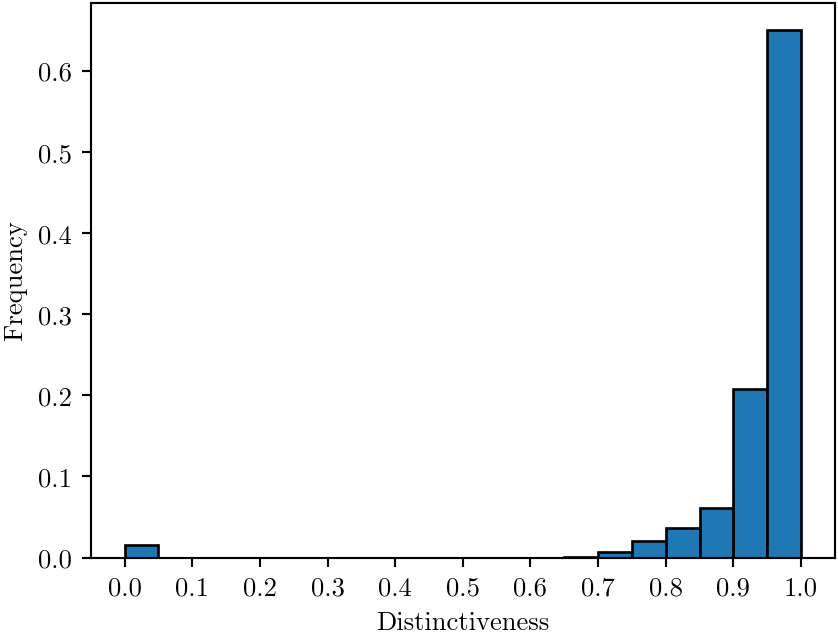}
    \caption{
      The histogram of the distinctiveness values of the ground truth LTL formulae,
      calculated using 1000 test set pairs from LTLRandom35.
    }
    \label{ground:distinc-hist}
  \end{minipage}%
  \hfill
  \begin{minipage}{.48\textwidth}
    \centering
    \includegraphics[width=\textwidth]{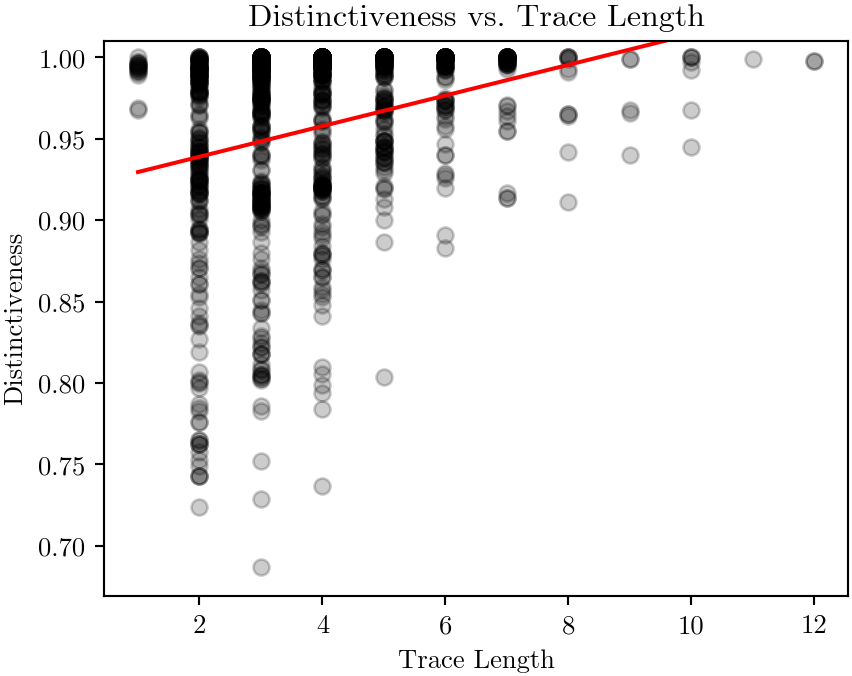}
    \caption{
      The scatter plot of the distinctiveness values reported in Fig.~\ref{ground:distinc-hist} vs. the sequence length of the input symbolic trace (total length of $u$ and $v$).
      The line of best fit is shown in red.
    }
    \label{ground:distinc-vs-tracelen}
  \end{minipage}
  \begin{minipage}{.48\textwidth}
    \centering
    \includegraphics[width=\textwidth]{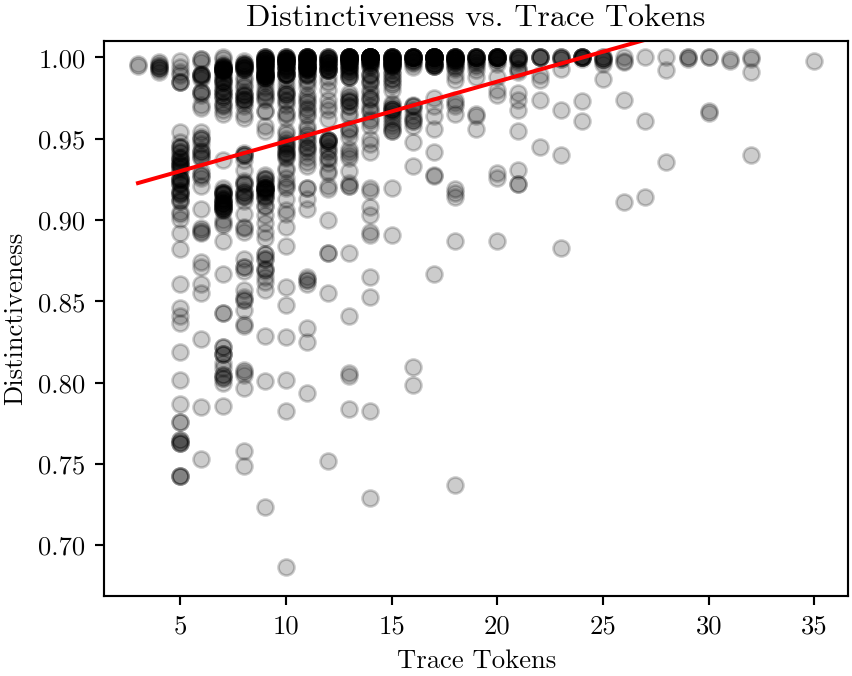}
    \caption{
      The scatter plot of the distinctiveness values reported in Fig.~\ref{ground:distinc-hist} vs. the number of tokens in the input symbolic trace.
      The line of best fit is shown in red.
    }
    \label{ground:distinc-vs-tracetokens}
  \end{minipage}%
  \hfill
  \begin{minipage}{.48\textwidth}
    \centering
    \includegraphics[width=\textwidth]{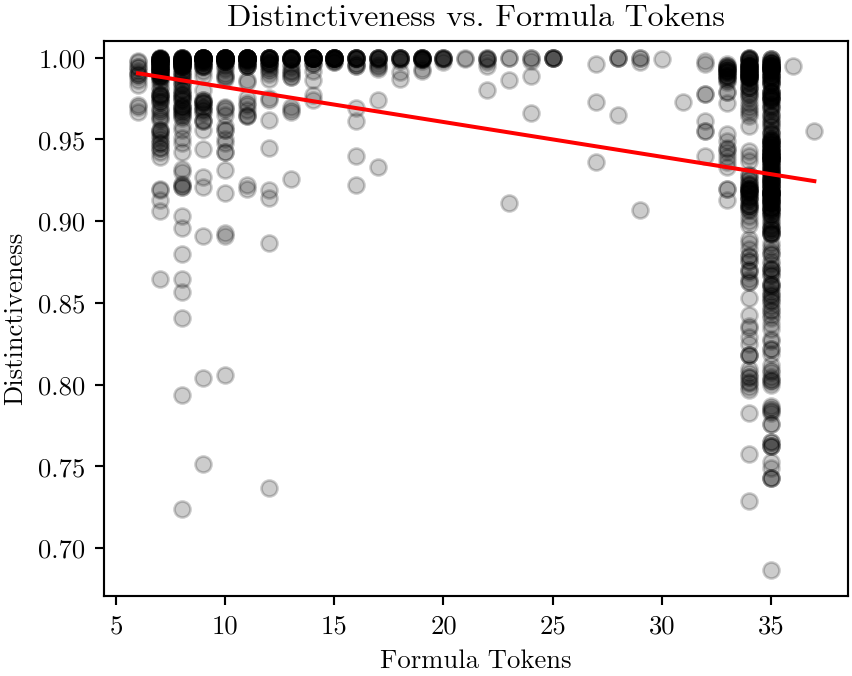}
    \caption{
      The scatter plot of the distinctiveness values reported in Fig.~\ref{ground:distinc-hist} vs. the number of tokens in the generated output.
      The line of best fit is shown in red.
    }
    \label{ground:distinc-vs-ltltokens}
  \end{minipage}
\end{figure}

In Table~\ref{test-distinc-full}, we give the full distinctiveness evaluation of the three training runs of the best models for each architecture, alongside the distinctiveness evaluation of the ground truth formulae for comparison.
Figures~\ref{mamba:distinc-vs-tracetokens} and \ref{mamba:distinc-vs-ltltokens} compare the distinctiveness values reported in Fig.~\ref{fig:distinc-hist} (Mamba 128-6-a) against the trace token count and the generated formula token count respectively.
We omit the corresponding plots for other models since the results were very similar.
Figures~\ref{ground:distinc-hist} to \ref{ground:distinc-vs-ltltokens} analyze the distinctiveness values of 1000 ground truth formulae from the LTLRandom35's test set.

\begin{table}[htbp]
  \caption{
    Distinctiveness evaluation on 1000 samples from the LTLRandom35's test set.
    ``Avg.'' column gives the average and the standard deviation.
    Columns ``Q1'' to ``Q3'' represent quartiles.
    ``Perfect'' represents the number of cases in which the distinctiveness value is $1.0$, the highest attainable value.
  }
  \label{test-distinc-full}
  \centering
  \begin{tabular}{llccccr}
    \toprule
    & & \multicolumn{5}{c}{Distinctiveness} \\
    \cmidrule(r){3-7}
    Architecture & Model Name & Avg. & Q1 & Q2 & Q3 & Perfect \\
    \midrule
    \multirow{3}{*}{Transformer}
    & 8-128-512-6-a & $0.952 \pm 0.125$ & $0.932$ & $0.989$ & $0.999$ & $ 223$ \\
    & 8-128-512-6-b & $0.936 \pm 0.171$ & $0.928$ & $0.992$ & $0.999$ & $ 219$ \\
    & 8-128-512-6-c & $0.948 \pm 0.129$ & $0.931$ & $0.989$ & $0.999$ & $ 217$ \\
    \midrule
    \multirow{3}{*}{Mamba}
    & 128-6-a       & $0.950 \pm 0.126$ & $0.930$ & $0.987$ & $0.999$ & $ 215$ \\
    & 128-6-b       & $0.951 \pm 0.126$ & $0.928$ & $0.989$ & $0.999$ & $ 217$ \\
    & 128-6-c       & $0.948 \pm 0.126$ & $0.921$ & $0.988$ & $0.999$ & $ 224$ \\
    \midrule
    \multirow{3}{*}{Llama}
    & 8-64-128-6-a  & $0.949 \pm 0.127$ & $0.926$ & $0.988$ & $0.999$ & $ 222$ \\
    & 8-64-128-6-b  & $0.950 \pm 0.128$ & $0.930$ & $0.989$ & $0.999$ & $ 223$ \\
    & 8-64-128-6-c  & $0.947 \pm 0.130$ & $0.932$ & $0.988$ & $0.999$ & $ 226$ \\
    \midrule
    \multicolumn{2}{l}{Ground Truth Formulae}
    & $0.942 \pm 0.129$ & $0.926$ & $0.981$ & $0.997$ & $ 147$ \\
    \bottomrule
  \end{tabular}
\end{table}

\section{The Effect of Syntax Enforcing}
\label{sec:enforcing-exp}

\begin{algorithm}[H]
    \caption{Enforce LTL syntax constraints}
    \label{alg:enforce}
    \begin{algorithmic}[1]
        \Function{Enforce}{$\text{input\_ids}: \text{torch.Tensor}, \text{logits}: \text{torch.Tensor}$}
            \State $\text{expected\_statements} \gets 1$ \Comment{The number of remaining statements.}
            \For{$\text{token}$ \textbf{in} $\text{input\_ids}$}
                \If{$\text{expected\_statements} = 0$}
                    \State \textbf{assert} $\text{token} = $ \texttt{EOS}
                    \State \textbf{break}
                \EndIf
                \State $\text{operand\_count} \gets \text{get\_operand\_count(token)}$
                \State $\text{expected\_statements} \gets \text{expected\_statements} + (\text{operand\_count} - 1)$
            \EndFor
            \If{$\text{expected\_statements} == 0$} 
                \State $\text{logits}[:-1] \gets -\infty$ \Comment{The only legal token is EOS, which is last in vocabulary.}
            \Else
                \State $\text{logits}[-1] \gets -\infty$ \Comment{More statements are expected, EOS token is illegal.}
            \EndIf
            \State \textbf{return} $\text{logits}$
        \EndFunction
    \end{algorithmic}
\end{algorithm}

The syntax enforcing algorithm is given in Algorithm~\ref{alg:enforce}. As previously explained in the paper, the algorithm avoids the generation of syntactically invalid formulae.

\subsection{On the test set}

Table~\ref{test-set-enforcing} shows the effect of syntax enforcing on the models we evaluated on the test set in Appendix~\ref{sec:full-test}.
In total, our syntax enforcing algorithm converted $77\%$ of the invalid formulae into semantically correct formulae.
However, different models vary wildly in terms of how much they benefit from syntax enforcing.
In particular, before syntax enforcing, the Mamba architecture displayed a significantly higher tendency to abide by the syntax rules (768 invalid formulae), whereas the transformer encoder-decoder and Llama struggled (15967 and 20078 invalid formulae respectively).
Furthermore, although the invalid-to-correct conversion ratio is $93\%$ for the Llama models, this figure is only $57\%$ for the transformer encoder-decoder models.
Interestingly, almost all ($99\%$) invalid formulae by the Llama 8-64-128-6-a model became correct after syntax enforcing.

\begin{table}[htbp]
  \caption{
    The effect of syntax enforcing on the invalid formulae generated by the three training runs of the best models for each architecture on LTLRandom35's test set.
  }
  \label{test-set-enforcing}
  \centering
  \begin{tabular}{llrrrrrr}
    \toprule
    \multicolumn{2}{c}{Model} & &
    \multicolumn{5}{c}{After Syntax Enforcing} \\
    \cmidrule(r){1-2}
    \cmidrule(r){4-8}
    Architecture & Model Name & Total Invalid & Correct & Exact & Incorrect & Invalid & Timeout \\
    \midrule

    \multirow{3}{*}{Transformer}
    & 8-128-512-6-a & 0 & 0 & 0 & 0 & 0 & 0 \\
    & 8-128-512-6-b & 13392 & 7739 & 0 & 4748 & 44 & 861 \\
    & 8-128-512-6-c & 2575 & 1385 & 0 & 1150 & 40 & 0 \\
    \cmidrule(r){3-8}
    \multicolumn{2}{r}{Subtotal} & 15967 & 9124 & 0 & 5898 & 84 & 861 \\
    \midrule

    \multirow{3}{*}{Mamba}
    & 128-6-a & 1 & 0 & 0 & 1 & 0 & 0 \\
    & 128-6-b & 228 & 6 & 0 & 221 & 0 & 1 \\
    & 128-6-c & 539 & 537 & 0 & 2 & 0 & 0 \\
    \cmidrule(r){3-8}
    \multicolumn{2}{r}{Subtotal} & 768 & 543 & 0 & 224 & 0 & 1 \\
    \midrule

    \multirow{3}{*}{Llama}
    & 8-64-128-6-a & 13923 & 13781 & 0 & 27 & 0 & 115 \\
    & 8-64-128-6-b & 2092 & 1027 & 0 & 1065 & 0 & 0 \\
    & 8-64-128-6-c & 4063 & 3879 & 0 & 17 & 0 & 167 \\
    \cmidrule(r){3-8}
    \multicolumn{2}{r}{Subtotal} & 20078 & 18687 & 0 & 1109 & 0 & 282 \\

    \midrule
    \multicolumn{2}{l}{GRAND TOTAL} & 36813 & 28354 & 0 & 7231 & 84 & 1144 \\

    \bottomrule
  \end{tabular}
\end{table}

\subsection{On the validation set, for all models}

Tables~\ref{ted-enforcing}, \ref{mamba-enforcing} and \ref{llama-enforcing} show how syntax enforcing changes the invalid formulae generated by the transformer encoder-decoder, Mamba models, and Llama-based models respectively.
This test was performed on the same 10000 validation set samples used in Appendix~\ref{sec:hyperparam}.
The models that didn't generate any invalid formulae are omitted from the tables.

Excluding the 32 dimensional transformer models which frequently fail to terminate their predictions,
21920 invalid formulae were generated in total across all architectures and models.
18325 ($83.6\%$) of these invalid formulae were converted into correct formulae after syntax enforcing.

Comparing the architectures, once more we observe that the Mamba models have a much lower tendency to generate syntactically invalid formulae when syntax enforcing is disabled.
Furthermore, all invalid formulae are eliminated by syntax enforcing in Mamba models in this experiment, which is not the case for other models.

\begin{table}[htbp]
  \caption{
    The effect of syntax enforcing on the invalid formulae generated by the transformer encoder-decoder models on 10000 validation set samples.
  }
  \label{ted-enforcing}
  \centering
  \begin{tabular}{lrrrrrrr}
    \toprule
    \multicolumn{2}{c}{Model} & &
    \multicolumn{5}{c}{After Syntax Enforcing} \\
    \cmidrule(r){1-2}
    \cmidrule(r){4-8}
    Name & Parameters & Total Invalid & Correct & Exact & Incorrect & Invalid & Timeout \\
    \midrule
    4-32-64-2 & 44,142 & 3038 & 157 & 0 & 4 & 2877 & 0 \\
    4-32-64-4 & 86,894 & 583 & 410 & 0 & 63 & 98 & 12 \\
    4-32-64-6 & 129,646 & 3817 & 691 & 0 & 21 & 3074 & 31 \\
    4-64-256-4 & 469,710 & 51 & 51 & 0 & 0 & 0 & 0 \\
    4-128-512-4 & 1,856,910 & 1 & 1 & 0 & 0 & 0 & 0 \\
    8-64-128-2 & 170,190 & 38 & 14 & 0 & 15 & 0 & 9 \\
    8-64-128-4 & 337,614 & 531 & 524 & 0 & 0 & 0 & 7 \\
    8-64-128-6 & 505,038 & 3653 & 3511 & 5 & 24 & 2 & 111 \\
    8-64-256-2 & 236,238 & 0 & 0 & 0 & 0 & 0 & 0 \\
    8-128-256-2 & 668,046 & 48 & 2 & 0 & 2 & 44 & 0 \\
    8-128-256-8 & 2,655,630 & 582 & 560 & 0 & 22 & 0 & 0 \\
    8-128-512-2 & 931,214 & 468 & 464 & 0 & 2 & 2 & 0 \\
    8-128-512-6-a & 2,782,606 & 0 & 0 & 0 & 0 & 0 & 0 \\
    8-128-512-6-b & 2,782,606 & 1362 & 820 & 0 & 462 & 4 & 76 \\
    8-128-512-6-c & 2,782,606 & 264 & 138 & 0 & 125 & 1 & 0 \\
    8-128-512-8 & 3,708,302 & 133 & 104 & 0 & 29 & 0 & 0 \\
    8-128-1024-8 & 5,813,646 & 932 & 527 & 0 & 404 & 1 & 0 \\
    \midrule
    \multicolumn{2}{l}{TOTAL} & 15501 & 7974 & 5 & 1173 & 6103 & 246 \\
    \bottomrule
  \end{tabular}
\end{table}

\begin{table}[htbp]
  \caption{
    The effect of syntax enforcing on the invalid formulae generated by the Mamba models on 10000 validation set samples.
  }
  \label{mamba-enforcing}
  \centering
  \begin{tabular}{lrrrrrrr}
    \toprule
    \multicolumn{2}{c}{Model} & &
    \multicolumn{5}{c}{After Syntax Enforcing} \\
    \cmidrule(r){1-2}
    \cmidrule(r){4-8}
    Name & Parameters & Total Invalid & Correct & Exact & Incorrect & Invalid & Timeout \\
    \midrule
    32-2 & 20,960 & 170 & 129 & 0 & 39 & 0 & 2 \\
    32-4 & 40,864 & 1 & 1 & 0 & 0 & 0 & 0 \\
    32-6-a & 60,768 & 0 & 0 & 0 & 0 & 0 & 0 \\
    32-6-b & 60,768 & 21 & 4 & 0 & 0 & 0 & 17 \\
    32-6-c & 60,768 & 6 & 6 & 0 & 0 & 0 & 0 \\
    32-6-d & 60,768 & 2 & 1 & 0 & 1 & 0 & 0 \\
    64-2 & 67,520 & 104 & 97 & 0 & 2 & 0 & 5 \\
    64-4 & 132,928 & 2 & 2 & 0 & 0 & 0 & 0 \\
    64-6 & 198,336 & 312 & 311 & 0 & 0 & 0 & 1 \\
    128-2 & 237,440 & 622 & 611 & 0 & 4 & 0 & 7 \\
    128-4 & 470,656 & 453 & 452 & 0 & 1 & 0 & 0 \\
    128-6-a & 703,872 & 0 & 0 & 0 & 0 & 0 & 0 \\
    128-6-b & 703,872 & 18 & 0 & 0 & 18 & 0 & 0 \\
    128-6-c & 703,872 & 46 & 46 & 0 & 0 & 0 & 0 \\
    256-2 & 884,480 & 12 & 1 & 0 & 11 & 0 & 0 \\
    256-4 & 1,760,512 & 1 & 1 & 0 & 0 & 0 & 0 \\
    256-6 & 2,636,544 & 299 & 258 & 0 & 41 & 0 & 0 \\
    512-2 & 3,407,360 & 142 & 20 & 0 & 122 & 0 & 0 \\
    512-4 & 6,797,824 & 31 & 23 & 0 & 8 & 0 & 0 \\
    512-6 & 10,188,288 & 20 & 4 & 0 & 16 & 0 & 0 \\
    \midrule
    \multicolumn{2}{l}{TOTAL} & 2262 & 1967 & 0 & 263 & 0 & 32 \\
    \bottomrule
  \end{tabular}
\end{table}

\begin{table}[htbp]
  \caption{
    The effect of syntax enforcing on the invalid formulae generated by the Llama-based models on 10000 validation set samples.
  }
  \label{llama-enforcing}
  \centering
  \begin{tabular}{lrrrrrrr}
    \toprule
    \multicolumn{2}{c}{Model} & &
    \multicolumn{5}{c}{After Syntax Enforcing} \\
    \cmidrule(r){1-2}
    \cmidrule(r){4-8}
    Name & Parameters & Total Invalid & Correct & Exact & Incorrect & Invalid & Timeout \\
    \midrule
    4-32-64-2 & 21,824 & 3 & 1 & 0 & 2 & 0 & 0 \\
    4-32-64-4 & 42,432 & 8 & 5 & 0 & 3 & 0 & 0 \\
    4-32-64-6 & 63,040 & 884 & 852 & 0 & 11 & 0 & 21 \\
    4-64-128-8 & 331,136 & 637 & 554 & 0 & 76 & 0 & 7 \\
    4-128-512-4 & 1,054,464 & 563 & 462 & 0 & 74 & 0 & 27 \\
    8-64-128-2 & 84,608 & 0 & 0 & 0 & 0 & 0 & 0 \\
    8-64-128-4 & 166,784 & 272 & 268 & 0 & 1 & 0 & 3 \\
    8-64-128-6-a & 248,960 & 1500 & 1486 & 0 & 7 & 0 & 7 \\
    8-64-128-6-b & 248,960 & 220 & 117 & 0 & 103 & 0 & 0 \\
    8-64-128-6-c & 248,960 & 431 & 404 & 0 & 1 & 0 & 26 \\
    8-64-128-8 & 331,136 & 877 & 857 & 0 & 4 & 0 & 16 \\
    8-128-256-2 & 333,056 & 282 & 252 & 0 & 25 & 0 & 5 \\
    8-128-256-8 & 1,317,632 & 1278 & 837 & 0 & 189 & 0 & 252 \\
    8-128-512-2 & 529,664 & 209 & 177 & 0 & 31 & 0 & 1 \\
    8-128-512-4 & 1,054,464 & 940 & 908 & 0 & 29 & 0 & 3 \\
    8-128-512-6 & 1,579,264 & 1306 & 1167 & 0 & 136 & 0 & 3 \\
    8-256-512-6 & 3,944,960 & 2185 & 1295 & 0 & 738 & 4 & 148 \\
    \midrule
    \multicolumn{2}{l}{TOTAL} & 11595 & 9642 & 0 & 1430 & 4 & 519 \\
    \bottomrule
  \end{tabular}
\end{table}

\end{document}